\documentclass[journal]{IEEEtran}
\usepackage[numbers,sort&compress]{natbib} 
\usepackage{graphicx,subfigure}
\usepackage{times}
\usepackage[linesnumbered,boxed,ruled,commentsnumbered]{algorithm2e}
\usepackage{graphicx}
\usepackage{amsmath,amssymb,amsthm}
\usepackage{hyperref}
\usepackage{booktabs}

\usepackage{bm}
\usepackage{makecell}
\usepackage{multirow}
\usepackage{color}
\usepackage{colortbl}
\usepackage{xcolor}

\ifCLASSINFOpdf
\else
\fi
\begin{document}
% paper title
% Titles are generally capitalized except for words such as a, an, and, as,
% at, but, by, for, in, nor, of, on, or, the, to and up, which are usually
% not capitalized unless they are the first or last word of the title.
% Linebreaks \\ can be used within to get better formatting as desired.
% Do not put math or special symbols in the title.
%\title{SUPERS: Suppressing Perturbing Sub-regions for Few-Shot Medical Image Segmentation}
\title{Partition-A-Medical-Image: Extracting Multiple Representative Sub-regions for Few-shot Medical Image Segmentation}
\author{Yazhou Zhu, Shidong Wang, Tong Xin, Zheng Zhang and Haofeng Zhang
\thanks{Manuscript received xxxx; revised xxxxx; accepted xxxxx. }
\thanks{This work was supported in part by the National Natural Science Foundation of China under Grant 62371235 and Grant 62072246, and in part by the Natural Science Foundation of Jiangsu Province under Grant BK20201306. (Corresponding author: Haofeng Zhang.) }
\thanks{Yazhou Zhu and Haofeng Zhang are with the School of Computer Science and Engineering, Nanjing University of Science and Technology, Nanjing, 210094, China. (e-mail: \{zyz\_nj, zhanghf\}@njust.edu.cn). }
\thanks{Shidong Wang is with the School of Engineering, Newcastle University, Newcastle upon Tyne, NE1 7RU, United Kingdom. (e-mail: shidong.wang@newcastle.ac.uk)}
\thanks{Tong Xin is with the School of Computing, Newcastle University, Newcastle upon Tyne, NE1 7RU, United Kingdom. (e-mail: tong.xin@newcastle.ac.uk)}
\thanks{Zheng Zhang is with the Shenzhen Key Laboratory of Visual Object Detection and Recognition, Harbin Institute of Technology, Shenzhen 518055, China. (e-mail: darrenzz219@gmail.com)}
}

\markboth{Preprint}%
{Shell \MakeLowercase{\textit{et al.}}: Bare Demo of IEEEtran.cls for IEEE Journals}

\maketitle

\begin{abstract}
Few-shot Medical Image Segmentation (FSMIS) is a more promising solution for medical image segmentation tasks where high-quality annotations are naturally scarce. However, current mainstream methods primarily focus on extracting holistic representations from support images with large intra-class variations in appearance and background, and encounter difficulties in adapting to query images. In this work, we present an approach to extract multiple representative sub-regions from a given support medical image, enabling fine-grained selection over the generated image regions. Specifically, the foreground of the support image is decomposed into distinct regions, which are subsequently used to derive region-level representations via a designed Regional Prototypical Learning (RPL) module. We then introduce a novel Prototypical Representation Debiasing (PRD) module based on a two-way elimination mechanism which suppresses the disturbance of regional representations by a self-support, Multi-direction Self-debiasing (MS) block, and a support-query, Interactive Debiasing (ID) block. Finally, an Assembled Prediction (AP) module is devised to balance and integrate predictions of multiple prototypical representations learned using stacked PRD modules. Results obtained through extensive experiments on three publicly accessible medical imaging datasets demonstrate consistent improvements over the leading FSMIS methods. The source code is available at \url{https://github.com/YazhouZhu19/PAMI}. 

\end{abstract}


\section{Introduction}
\label{sec:introduction}
\IEEEPARstart{M}{edical} image segmentation \cite{menze2014multimodal} aims to identify surface properties or volume of specific anatomical structures in various medical images, including X-ray, Ultrasonography, PET/CT, and MRI scans. Deep learning-based algorithms \cite{lecun2015deep,he2016deep} are particularly adept at this task because they can generate measurements and segments from medical images without the time-consuming manual work required by traditional methods \cite{9944699}. The effectiveness of deep learning algorithms depends heavily on the availability of large-scale, high-quality data that is fully annotated in a pixel-wise manner, which is naturally scarce in the field of medical image computing. Therefore, how to build a deep learning algorithm to effectively segment medical images using only a limited amount of labelled data is a critical yet challenging task.

To tackle this challenge, Few-Shot Learning (FSL) \cite{koch2015siamese,snell2017prototypical} is introduced to enable deep learning algorithms to extract useful knowledge when annotations are scarce. Formally, standard FSL algorithms first extract representative information from a small set of annotated data (\textit{Support Set}), and then ensure that the learned knowledge can be generalised to larger unannotated data (\textit{Query Set}). In general, there are three common strategies in FSL scenarios: meta-learning \cite{finn2017model}, prototypical network \cite{snell2017prototypical} and matching network \cite{vinyals2016matching}. Meta-learning methods aim to learn a good initialisation or set of parameters that allow models to quickly adapt to new tasks while both prototype and matching networks focus on extracting semantic or correlated information between support and query images. 

Analogues to the aforementioned FSL, Few-shot Semantic Segmentation (FSS) \cite{dong2018few,wang2019panet} mainly follows the idea based on PrototypicalNet \cite{snell2017prototypical} and MatchingNet \cite{lu2021simpler}, both of which aim to extract and aggregate class-specific semantic representations for fast adaptation from support to query. Specifically, PrototypicalNet-based methods learn to generate the key prototypical representation by applying Masked Average Pooling (MAP) on features extracted from support and query features, such as via clustering algorithms \cite{liu2020part,yang2020prototype} and attention mechanisms \cite{zhang2019pyramid,wang2020few}, and then refine prototypes by incorporating additional information from the background \cite{fan2022self,lang2022learning,liu2022learning}. MatchingNet-based methods seek how to establish robust associations between support and query features \cite{lu2021simpler,min2021hypercorrelation}. They might also incorporate intricate pixel-wise interactions between images and masks \cite{peng2023hierarchical} to further enhance their performance. The core of these MatchingNet-based strategies lies in extracting dense correspondences between query images and their corresponding support annotations. This process, In turn, greatly contributes to enhancing generalisation ability from support to query.

% \begin{figure}[!t]                   % htbp
% \centering
% \includegraphics[width=1\columnwidth]{fig2.pdf}
% \vspace{-3ex}
% \caption{Overview of previous few-shot medical image segmentation methods: (a) the two-branch interaction-based method, (b) the prototypical network-based method.}
% \vspace{-1ex}
% \label{Comparison}
% \end{figure}

\begin{figure*}[ht]                   % htbp
\centering
\includegraphics[width=0.99\textwidth]{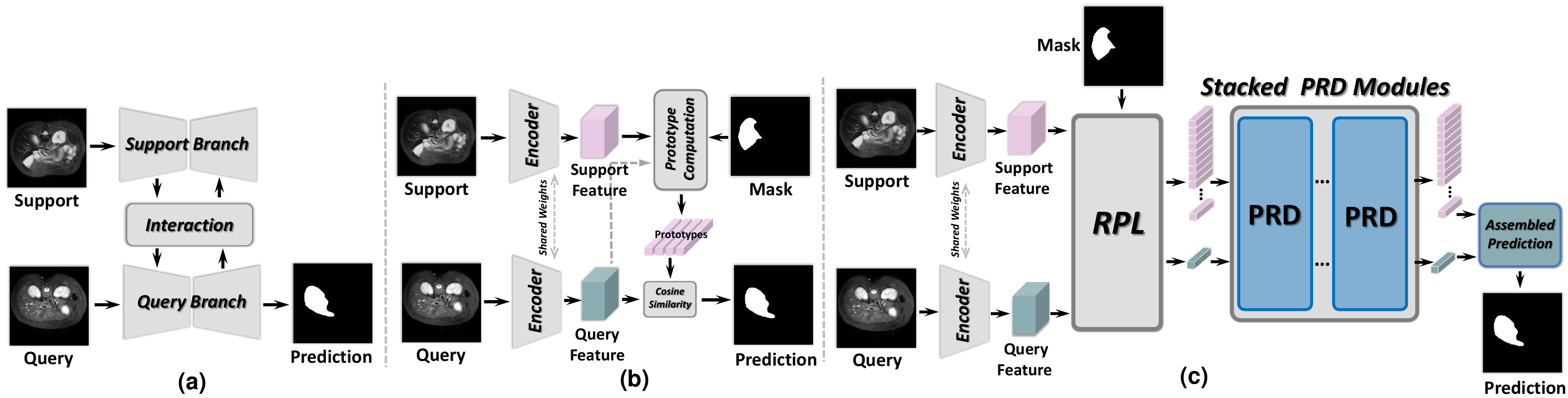}
\vspace{-1ex}
\caption{Comparison between previous few-shot medical image segmentation methods and our proposed method: (a) two-branch interaction-based method, (b) prototypical network-based method, (c) our proposed method which consists of a weights-shared feature encoder,  a Regional Prototypical Learning (RPL) module, a stacked of Prototypical Representation Debiasing (PRD) modules and an Assembled Prediction (AP) module.}
\vspace{-2ex}
\label{Comparison}
\end{figure*}

Given the advantages of few-shot learning in terms of sample size requirements, it has been naturally introduced into the field of medical image processing which is known as Few-Shot Medical Image Segmentation (FSMIS) \cite{ouyang2022self}. Existing algorithms in this domain \cite{roy2020squeeze,feng2021interactive,sun2022few,wu2022dual,ding2023few,tang2021recurrent,ouyang2022self,hansen2022anomaly,wang2022few} can be broadly classified into two categories: interactive methods derived from SENet \cite{roy2020squeeze} (shown in Fig. \ref{Comparison}(a)) and prototypical network-based methods \cite{tang2021recurrent,ouyang2022self,hansen2022anomaly,wang2022few} (shown in Fig. \ref{Comparison}(b)). The key to leading the success of the interaction-based approach is the use of non-local attention mechanism \cite{wang2018non,feng2021interactive} and contrastive learning \cite{khosla2020supervised,wu2022dual} to work in parallel between the support and query arms in an interactive manner. Prototypical network-based approaches have emerged as dominant methods in FSMIS research. The core idea of some prominent examples like SSL-ALPNet \cite{ouyang2022self}, ADNet \cite{hansen2022anomaly}, and SR\&CL \cite{wang2022few} is to obtain semantic-level prototypes by compressing support features and subsequently produce predictions by matching them with query features. 

Despite the success of the above methods, they fail to address the problem of the large \textbf{intra-class variations} posed by significant intra-class variations resulting from the inherent diversity of a specific organ, including \textit{size}, \textit{shape} and \textit{contour} which can vary across different patients or under distinct acquisition protocols. In particular, a certain number of disparate regions (identified as \textit{perturbing} regions) arise between the support images and query images which have the potential to degrade the generalisation capability of the obtained prototypes to some extent. The prototypes generated by the conventional simple masked average pooling (MAP) operation, therefore, are consistently inefficient and imprecise for the FSMIS task. 

\begin{figure}[!t]                   % htbp
\centering
\includegraphics[width=1\columnwidth]{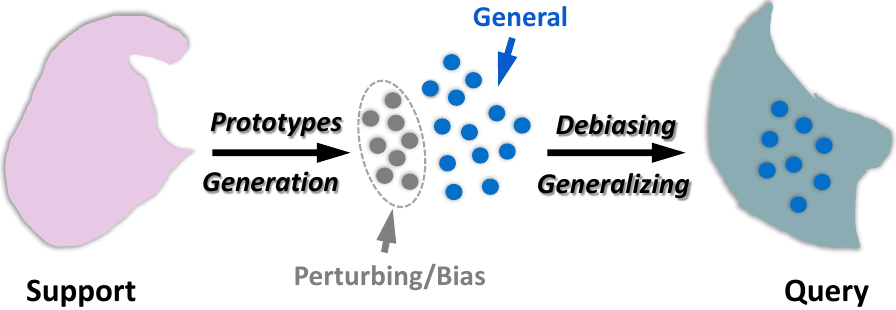}
\vspace{-1ex}
\caption{Motivation: The objective is to eliminate the perturbing elements in the partial prototypical representations within multiple foreground support prototypes while preserving and refining the remaining general prototypical representations for rapid generalization to query images. }
\label{motivation}
\end{figure}

To cope with this challenge, as depicted in Fig. \ref{motivation}, we introduce a new concept, Partition-A-Medical-Image (PAMI), which aims to learn multiple precise sub-region representations by partitioning a medical image into multiple sub-regions and mitigating the impact of the \textit{perturbing} sub-regions at the prototypical representation level, while refining the remaining areas. Concretely, it presents a Regional Prototypical
Learning (RPL) module to first strip the perturbing regions from the support foreground using a Voronoi-based method \cite{aurenhammer1991voronoi,zhang2022feature}, and then generate multiple separated regional-level prototypical representations. We also introduce the Prototypical Representation Debiasing (PRD) module, which employs two elimination methods: a self-support approach and a support-query interactive approach, capable of re-weighting region-level prototypical representations while selecting non-perturbing prototypical representations. By stacking multiple PRD modules, it can produce debiased prototypical representations for both supports and queries, which will be assembled and used for prediction. Our contributions are summarised as follows:  
%to obtain the accurate prototypical representations for FSMIS task. Driven by this motivation, we propose a novel method called Partition-A-Medical-Image (PAMI): Extracting Multiple Representative Sub-Regions for Few-Shot Medical Image Segmentation. %As shown in Fig. \ref{Overview of our method}, our proposed method consists of four main components: (a) a weights-shared feature extractor, (b) a Regional Prototypical Representations Generation (RPRG) module, (c) stacked Prototypical Representation Debiasing (PRD) modules and (d) an Assembled Prediction (AP) module. 

\begin{itemize}
    \item We introduce the new method, Partition-A-Medical-Image (PAMI) to alleviate the effects of intra-class variations by suppressing perturbations of regional prototypes. 
    %\item A strategy of suppressing perturbing regional prototypical representations to tackle with the intra-class problem in few-shot medical image segmentation task is proposed.
    \item A Regional Prototypical Learning (RPL) module is designed to derive multiple regional prototypes for the support and the coarse query prototype. 
    %\item A Regional Prototypical Representations Generation (RPRG) module is proposed. RPRG partitions the entire foreground using a Voronoi-based method and acquire multiple regional support prototypical representations for next step prototypical representation debiasing. The Query Prototype Generation (QPG) block is also designed in RPRG to provide coarse query prototype for enhancing the support prototypical representations.  
    \item A Prototypical Representation Debiasing (PRD) module is proposed to remove the biases of regional prototypical representations under the synergistic work of a self-support, Multi-direction Self-debiasing (MS) block and a support-query, Interactive Debiasing (ID) block.     
    %\item Stacked Prototypical Representation Debiasing (PRD) modules are employed to perform debiasing in two manners, including self-support manner by Multi-Direction Self-Debiasing (MDSD) and support-query interactive manner by Interactive Debiasing (ID), to obtain the ideal prototypical representations for query prediction. 
    %\item An Assembled Prediction (AP) module is proposed to balance and integrate predictions given by debiased support prototype and generated query prototype. 
    \item The proposed PAMI method can achieve state-of-the-art performance on three experimental datasets commonly used in medical image segmentation tasks.
\end{itemize}

\section{Related Work}
\subsection{Medical Image Segmentation}

Medical image segmentation \cite{9740186,9940997,10138036} constitutes a vital and foundational technique in numerous clinical research endeavours and practical applications. In recent years, methods based on deep learning have showcased unparalleled performance across an extensive array of medical image segmentation tasks, encompassing diverse areas such as tissues, organs, lesions, and tumours. The most acclaimed method is U-Net \cite{ronneberger2015u}, which employs an encoder-decoder architecture combined with specialised skip connections, facilitating advanced high-level structure extraction while maintaining texture fidelity in segmentation tasks. Subsequent to the development of U-Net, various modified versions have been proposed to further augment performance. Examples include U-Net++ \cite{zhou2019unet++} and nnUNet \cite{isensee2021nnu}, which place emphasis on optimizing internal skip connections and granting increased consideration to data preprocessing pipelines. Additionally, the integration of self-attention mechanisms within the domain has yielded notable results, as evidenced by works such as vanilla Transformer based methods \cite{9785614,chen2021transunet}, Swin-UNet \cite{cao2023swin}, TransBTS \cite{wang2021transbts}, and TransFuse \cite{zhang2021transfuse}. These approaches have demonstrated exceptional outcomes on a series of publicly accessible medical imaging datasets. However, it is important to note that the effectiveness of contemporary medical image segmentation methods remains heavily contingent upon the presence of copious manual annotations. This dependency may potentially impose constraints on the applicability of these methods within real-world clinical settings.

\subsection{Few-Shot Semantic Segmentation}

The field of FSS has emerged as an innovative solution to address the scarcity of annotated data in semantic segmentation. Currently, FSS methodologies can be broadly delineated into two primary categories: PrototypicalNet-based approaches and MatchingNet-based approaches. PrototypicalNet-based methods primarily focus on constructing accurate and generalizable prototypical representations derived from the extracted support and query features. To achieve this objective, several cutting-edge FSS strategies suggest aggregating multiple prototypical representations at the pixel-level or region-level for distinct semantic classes by employing advanced methodologies, including clustering \cite{yang2021mining,li2021adaptive}, Expectation-Maximization (EM) \cite{yang2020prototype}, transformers \cite{zhang2022mask,zhang2021few}, and others. MatchingNet-based FSS approaches aim to resolve the issue by establishing correspondences between support and query features \cite{min2021hypercorrelation} and implementing a pixel-wise interaction mechanism between images and masks \cite{peng2023hierarchical}. The foundation of MatchingNet-based approaches is to extract dense correspondences between query images and support annotations, ultimately aiming to enhance the method's generalization capabilities. Moreover, contemporary research efforts have reevaluated the FSS task from novel research perspectives, such as leveraging knowledge from non-target regions to bolster generalization ability \cite{liu2022learning,lang2022learning,lang2022beyond} and re-engineering the feature extractor while addressing various FSS-related challenges \cite{hu2023suppressing}.

\begin{figure*}[ht]                   % htbp
\centering
\includegraphics[width=0.98\textwidth]{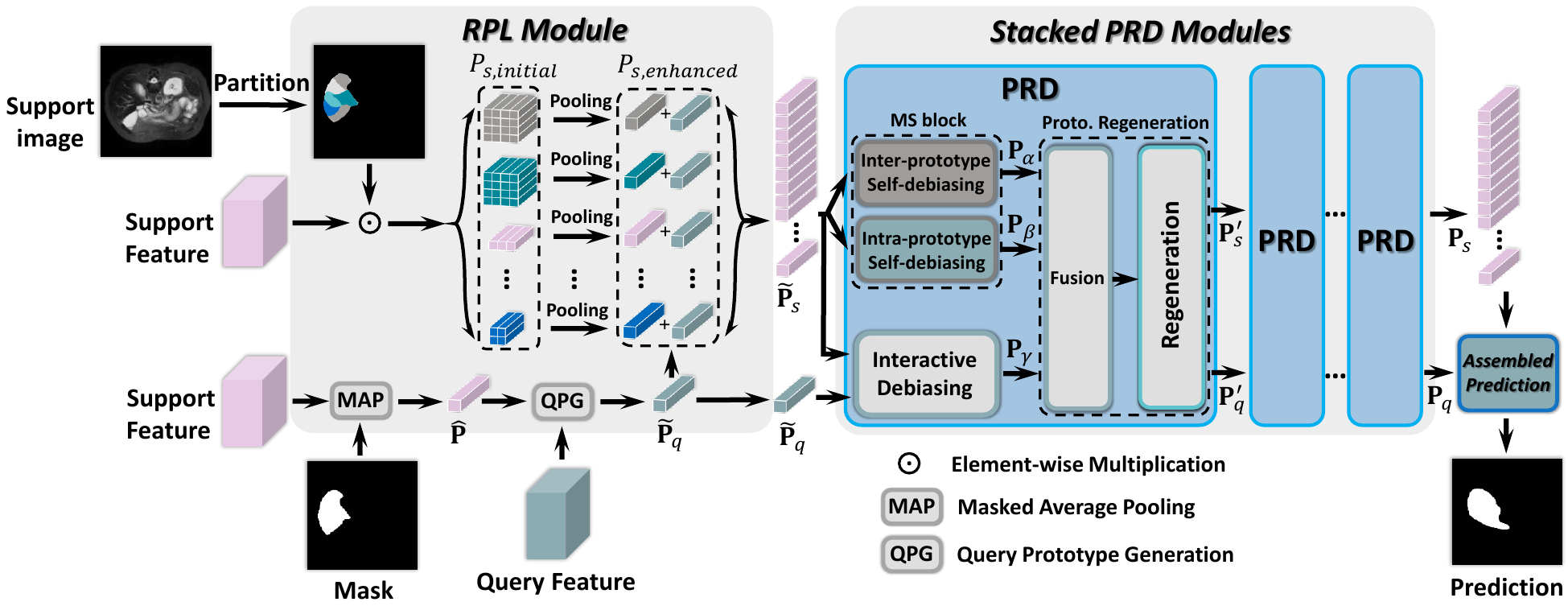}
\vspace{-1ex}
\caption{Details of the proposed Region Prototype Learning (RPL) module and the internal structures of stacked Prototypical Representation Debiasing (PRD) modules. The RPL module contains two parallel calculating pathways: the regional prototypes computation branch (top), and the coarse prototype computation branch (bottom). Each PRD module has three functional blocks: a Multi-direction Self-debiasing (MS) block, an Interactive Debiasing (ID) block and a Prototype Regeneration (PR) block.}
\vspace{-2ex}
\label{The core of our method}
\end{figure*}

\subsection{Few-Shot Medical Image Segmentation}

In the field of medical imaging, FSMIS has garnered substantial interest among researchers due to the practical challenges associated with accessing large-scale medical imaging datasets, considering legal, ethical, and user privacy constraints. FSMIS methods can be classified into two primary research streams: those employing a two-branch interactive structure \cite{roy2020squeeze,feng2021interactive,sun2022few,wu2022dual,ding2023few} and those based on prototypical network structures \cite{tang2021recurrent,ouyang2022self,hansen2022anomaly,wang2022few}. Distinct characteristics of medical images, as opposed to natural images, necessitate the development of unique algorithmic approaches for processing medical images compared to those used in natural image processing. For instance, the diverse and heterogeneous textures inherent in medical images warrant greater emphasis on extracting general and generalizable features, while the considerable variance in image intensity requires methods with enhanced discriminatory capabilities. Specific challenges emerge in few-shot medical image segmentation when transferring medical support images to medical query images. The first category of FSMIS methods focuses on constructing innovative connections and interactions between support and query images using novel mechanisms, considering that organs and lesions typically occupy specific locations within the abdomen and tumors. A common practice among these FSMIS methods incorporates various attention mechanisms into the interaction block. Approaches such as SE-Net \cite{roy2020squeeze}, MRrNet \cite{feng2021interactive}, and GCN-DE \cite{sun2022few} combine attention mechanism variants with specialized architectures tailored for medical scenarios. Additionally, AAS-DCL \cite{wu2022dual} proposes implementing contrastive learning among different prototypes to elevate the performance of FSMIS methods to new heights. The second category of FSMIS methods primarily adheres to the technical principles of classical prototypical network-based FSS methods. A representative work, SSL-ALPNet \cite{ouyang2022self}, introduces a novel Adaptive Local Prototype pooling module (ALP) designed to augment the generalization ability of prototype representations by extracting localized object information.

\section{Methodology}
\subsection{Problem Settings}
The goal of FSMIS is to accurately segment the object of unseen classes $\mathcal{C}_{novel}$ by using a limited number of well-annotated images of known classes $\mathcal{C}_{known}$ from the base dataset $\mathcal{D}_{base}$, where $\mathcal{C}_{known} \cap \mathcal{C}_{novel} = \emptyset$. Concretely, $\mathcal{D}_{base}$ contains a collection of image-mask pairs, expressed by: $(\textbf{I}^{j}, \mathcal{M}^{j})_{j=1}^{N}$, in which $\mathcal{M}^{j}$ is the semantic mask for the training image $\textbf{I}^{j}$, and $N$ represents the total number of image-mask pairs. In the testing phase, the support image set $S=(\textbf{I}^{i}_{s}, \mathcal{M}^{i}_{s})^{k}_{i=1} \in \mathcal{C}_{novel}$ is introduced into the task, where $\textbf{I}^{i}_{s}$ denotes the support image and $\mathcal{M}^{i}_{s}$ is the corresponding mask for foreground object in $\textbf{I}^{i}_{s}$, and $k$ is the number of image-mask pairs within the support set which is usually set as 1 for 1-shot or 5 for 5-shot. The evaluation of FSMIS methods is conducted on the query set: $Q = (\textbf{I}_q, \mathcal{M}_q) \in \mathcal{C}_{novel}$, where $\textbf{I}_q$ denotes the query image and $\mathcal{M}_q$ is the corresponding ground-truth mask. In essence, FSMIS methods employ the support set $S$ to generate a predicted segmentation mask $\tilde{\mathcal{M}}_{f}$ for each image $\textbf{I}_{q}$ within the query set $Q$.

\subsection{Architecture Overview}
The overall workflow of the proposed method is depicted in Fig. \ref{Comparison}(c), which contains four key components: (a) weights-shared encoder for feature extraction $\textbf{F} = f_{\theta}(\textbf{I})$, where $\theta$ denotes the model parameters; (b) a Regional Prototypical Learning (RPL) module for disassembling foreground region and bringing in auxiliary query information; (c) a stacked of multiple Prototypical Representation Debiasing (PRD) modules for prototypical feature debiasing, fusion and regeneration; (d) An Assembled Prediction (AP) module for the query mask prediction. Details of the proposed framework are illustrated in Fig. \ref{The core of our method}. Following the experimental settings in \cite{hansen2022anomaly}, the ResNet101 \cite{he2016deep} is chosen as the backbone of $f_{\theta}$ which has been pre-trained on the MS-COCO dataset. Then, the feature $\textbf{F}_{s}$ and $\textbf{F}_{q}$ extracted from support image and query image by extractor $f_{\theta}$ are fed into the RPL module for generating multiple regional prototypical representations $P_{s, enhanced}$, which will be rectified and debiased by following stacked PRD modules to produce the optimal prototypical representation. Notably, the input data is first processed using a 3D-based supervoxel clustering algorithm \cite{hansen2022anomaly,ouyang2022self} to generate pseudo-masks, where the generated masks are treated as supervision to later implement few-shot learning in a meta-learning-based episodic training manner. 

%Additionally, here it is necessary to briefly explain how data is processed before introducing the details of our proposed method. Firstly, 3D based supervoxel clustering preprocessing algorithm \cite{hansen2022anomaly,ouyang2022self} is employed to generate pseudo-masks as supervision for totally self-supervised learning manner in training phrase, which makes method training get rid of any manual annotations. Then, meta-learning based episodic training manner is also constructed with above receiving pseudo-masks to implement few-shot learning protocol. 

\subsection{Regional Prototype Learning} \label{sec3c}
The Regional Prototype Learning (RPL) module is proposed based on the fact that not all sub-regions extracted according to the support foreground are firmly related to the query image. To this end, we attempt to partition the support foreground using the Voronoi-based method to produce multiple region prototypes, where the perturbation information in the corresponding prototypical representations can be debiased by subsequent operations.    
%Considering some sub-regions of support foreground are not transferable for query image, in this paper, after obtaining support and query features $\textbf{F}_{s}, \textbf{F}_{q}$ by using extractor $f_{\theta}$, we introduce the Regional Prototype Learning (RPL) module to first obtain the partitioned sub-regions of support foreground with Voronoi-based method and then generate corresponding a set of multiple regional prototypical representations. With these prototypical representations, we can conduct a series of operations (such as debiasing) to suppress perturbing information and refine for the optimal prototypical representation.  

\begin{figure}[!t]                   % htbp
\centering
\includegraphics[width=0.98\columnwidth]{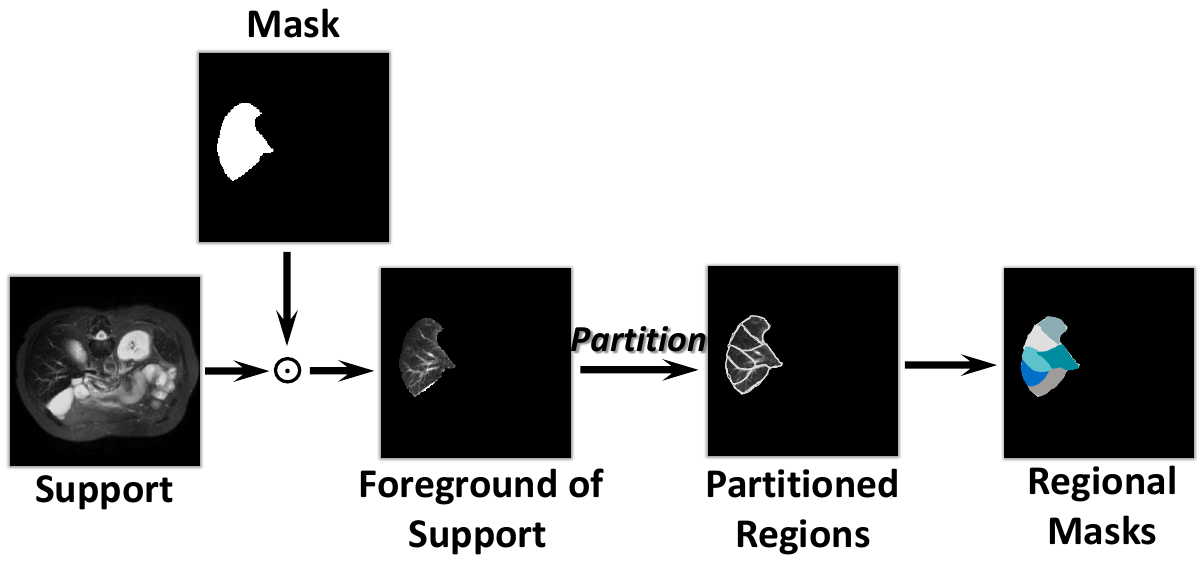}
\vspace{-1ex}
\caption{Workflow for generating region masks. The support foreground region is computed by multiplying the support image with its mask. Next, the partitioned regions of the support image can be obtained by performing a Voronoi-based method on the foreground region and then resulting in the region masks. }\centering
\vspace{-3ex}
\label{Regional masks generation workflow}
\end{figure}

As shown in Fig. \ref{The core of our method}, the RPL module consists of two branches computing in parallel: the regional prototypes computation branch and the coarse prototype computation branch. Concretely, the regional prototypes computation branch responds to produce the foreground region of the support image $R_f$ by taking the product of a given support image $\textbf{I}_{s} \in \mathbb{R}^{H \times W}$ and the corresponding foreground mask $\mathcal{M}^{f} \in \mathbb{R}^{H \times W}$. The generated foreground region will be partitioned by using the Voronoi-based method \cite{aurenhammer1991voronoi,zhang2022feature} to produce $N_{f}$ partitioned regions $S = \left \{ \mathcal{R}_n \right \}^{N_{f}}_{n=1} $ and a set of regional masks $\left \{ \mathcal{V}_n \right \}^{N_{f}}_{n=1}, \mathcal{V}_n \in \mathbb{R}^{H \times W} $, where $N_f$ is set as 64 in our method. The workflow of this step is illustrated in Fig. \ref{Regional masks generation workflow}, and the details of how to determine this value will be depicted in Section \ref{sec:exp}. With the available region masks $\left \{ \mathcal{V}_n \right \}^{N_{f}}_{n=1}$, we can determine that the set of the initial prototypical representations of the regional support $ P_{s,initial} = \left \{  \textbf{P}_{n} \right \}^{N_{f}}_{n=1}, \textbf{P}_{n} \in \mathbb{R}^{1 \times C}$ by:
\begin{equation}
\label{eq3}
\textbf{P}_{n} = \operatorname{MAP}(\textbf{F}_{s}, \mathcal{V}_{n}) = \frac{1}{\left |  \mathcal{V}_n\right | } \sum_{i=1}^{HW}\textbf{F}_{s,i}\mathcal{V}_{n,i},
\end{equation}
where $\textbf{P}_{n}$ denotes one regional support prototypical representation, $\textbf{F}_{s}\in \mathbb{R}^{C \times h \times w}$ is the support feature extracted by the encoder $f_{\theta}$ and $\textbf{F}_{s}$ is up-sampled into shape $(C,H,W)$, $\mathcal{V}_{n,i}$ denotes the $i_{th}$ regional mask and $\operatorname{MAP}(\cdot )$ is the masked average pooling operation. 

The coarse query prototype computation branch operates in parallel with the regional prototypes computation branch in order to enhance the information capability of the query image. Specifically, it first calculates the intermediate support prototype $\hat{\textbf{P}} \in \mathbb{R}^{1 \times C} $ on the support feature: $\hat{\textbf{P}}=\operatorname{MAP}(\textbf{F}_{s},\mathcal{M}^{f})$ using the MAP, and then produces the coarse query prototype $\tilde{\textbf{P}}_{q} \in \mathbb{R}^{1 \times C}$ by feeding $\hat{\textbf{P}}$ into the Query Prototype Generation (QPG) module. 

\begin{figure}[!t]                   % htbp
\centering
\includegraphics[width=0.78\columnwidth]{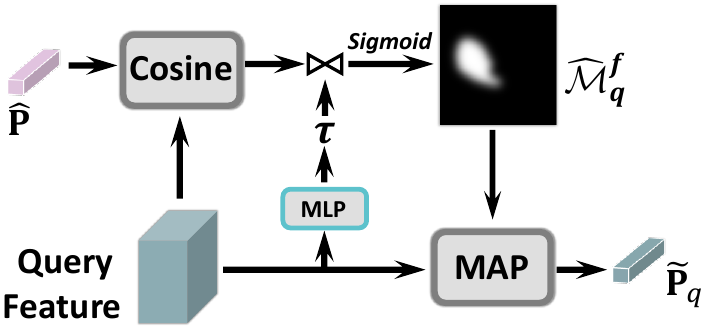}
\vspace{-1ex}
\caption{The Query Prototype Generation (QPG) block. }
\vspace{-2ex}
\label{QPG block}
\end{figure}

As illustrated in Fig. \ref{QPG block}, the coarse query foreground mask $\widehat{\mathcal{M}}_{q}^{f}$ is first calculated by: 
\begin{equation}
\label{QPC}
    \widehat{\mathcal{M}}_{q}^{f} = 1 - \sigma (S(\textbf{F}_{q}, \hat{\textbf{P}}) - \tau),
\end{equation} 
where $\textbf{F}_q \in \mathbb{R}^{C \times h \times w}$ is the feature extracted from the query image $\textbf{I}_q \in \mathbb{R}^{H \times W}$ by using the feature extractor $f_{\theta}$, $S(a, b) = -\alpha cos(a, b)$ is the negative cosine similarity with a fixed scaling factor $\alpha = 20$, $\sigma$ denotes the \textit{Sigmoid} activation function, and $\tau$ denotes a learnable threshold which can be derived by applying a single average pooling operation and a function $\operatorname{FC}(\cdot)$ containing two fully-connected layers to the query feature, expressed as $\tau = \operatorname{FC}(\textbf{F}_{q})$. As a consequence of this, the coarse query prototype $\tilde{\textbf{P}}_{q}  \in \mathbb{R}^{1 \times C}$ can be obtained by the masked average pooling operation, written as: 
\begin{equation}
\label{eq5}
\tilde{\textbf{P}}_{q} = \operatorname{MAP}(\textbf{F}_{q}, \widehat{\mathcal{M}}_{q}^{f}).
\end{equation}

Moreover, we also merge the coarse query prototype and obtained support prototypical representations to aggregate the query and the support information, which will lead to a set of enhanced prototypical representations of the query $P_{s,enhanced} = \left \{  \textbf{P}_{n}^{*} \right \}^{N}_{n=1}$, where $\textbf{P}_{n}^{*} = \textbf{P}_{n} +  \tilde{\textbf{P}}_{q}$, $\textbf{P}_{n}^{*} \in \mathbb{R}^{1 \times C}$.

\subsection{Stacked Prototypical Representation Debiasing Modules}
The purpose of introducing the stacked Prototypical Representation Debiasing (PRD) Modules is to filter the perturbation information present in $P_{s,enhanced}$ to the greatest extent. The number of PRD modules, $M$ is experimentally set to 5 for optimal performance (see Section \ref{sec:exp} for details). Each PRD module contains three key components, including the Multi-direction Self-debiasing (MS) block, the Interactive Debiasing (ID) block, and the Prototype Regeneration (PR) block. Taking the first PRD module as an example, the MS block takes the support prototypes $P_{s, enhanced}$ as the input to achieve the debiased prototype representations $\textbf{P}_{\alpha}, \textbf{P}_{ \beta}$ in a self-biasing manner including inter-prototype debiasing and intra-prototype debiasing. Meanwhile, the coarse query prototype $\tilde{\textbf{P}}_{q}$ is fed into the ID block together with $P_{s, enhanced}$ to calculate the affinity map which can realise prototype representation debiasing based on the self-selection mechanism, and result in the representation $\textbf{P}_{\gamma}$. Afterwards, the PR block fuses the representations of $\textbf{P}_{\alpha}, \textbf{P}_{\beta}$ and $\textbf{P}_{\gamma}$ and regenerates them into new prototypical presentations $\textbf{P}_{s}^{'}$ and $\textbf{P}_{q}^{'}$ as input to the next PRD module.

\begin{figure}[!t]                   % htbp
\centering
\includegraphics[width=1\columnwidth]{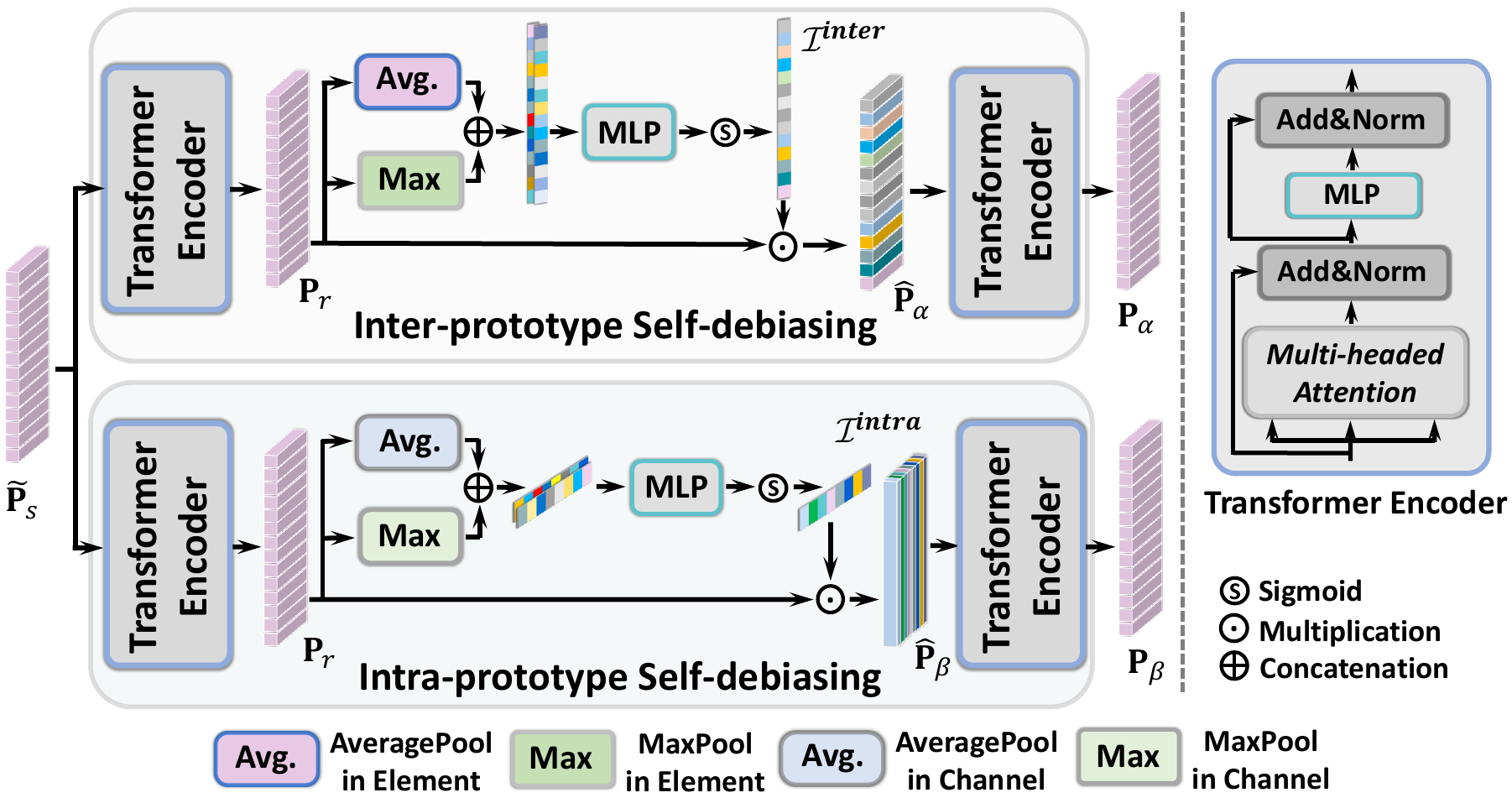}
\vspace{-1ex}
\caption{The Multi-direction Self-debiasing (MS) block is a two-way, self-debiasing block: an inter-prototype way and an intra-prototype way. Through these two self-debiasing methods, we can finally obtain the debiased prototypical representations $\textbf{P}_{\alpha}$ and $\textbf{P}_{\beta}$, respectively.}
\vspace{-2ex}
\label{MDSD block}
\end{figure}

\subsubsection{Multi-direction Self-debiasing}
The input to the MS block is the support prototypical representation $\tilde{\textbf{P}}_s \in \mathbb{R}^{N \times C}$, which is obtained by reshaping and rearranging all elements in $P_{s, enhanced}$. In particular, the support prototypical representation $\tilde{\textbf{P}}_s$ can be re-weighted through two-way debiasing operations, which are the inter-prototype (element dimension) and intra-prototype (channel dimension).    
%Before processed by MDSD, all elements in set $P_{s, enhanced}$ are first concatenated together into a support prototypical representation $\tilde{\textbf{P}}_s \in \mathbb{R}^{N \times C}$. In order to conduct self-support manner debiasing, as illustrated in the Fig. \ref{MDSD block}, we propose the Multi-Direction Self-Debiasing (MDSD) block to re-weight the support prototypical representation $\tilde{\textbf{P}}_s$ in inter-prototype (element dimension) and intra-prototype (channel dimension). 

In the inter-prototype self-debiasing pathway, the given $\tilde{\textbf{P}}_{s}$ is initially reconstructed into a representation $\textbf{P}_r \in \mathbb{R}^{N \times C}$ using a vanilla transformer encoder \cite{vaswani2017attention}, $\textbf{P}_r = \operatorname{E}_t(\tilde{\textbf{P}}_{s})$, where $\operatorname{E}_t$ consists of two sub-blocks: a multi-head attention (MHA) block and a multilayer perceptron (MLP) block. Formally, this process is denoted by: 
%In the inter-prototype self-debiasing pathway, the given $\tilde{\textbf{P}}_{s}$ is first reconstructed into representation $\textbf{P}_r \in \mathbb{R}^{N \times C}$  with using a vanilla transformer encoder \cite{vaswani2017attention} $\textbf{P}_r = \operatorname{E}_t(\tilde{\textbf{P}}_{s})$, where $\operatorname{E}_t$ consists of two sub-blocks including a multi-head attention (MHA) block and a multi-layer perceptron (MLP) block. Accordingly, the reconstruction process is written as follows: 
\begin{equation}
\label{eq6}
\begin{aligned} 
\textbf{P}^{i} =& \operatorname{LN}(\operatorname{MHA}(\tilde{\textbf{P}}_{s}, \tilde{\textbf{P}}_{s}, \tilde{\textbf{P}}_{s}) + \tilde{\textbf{P}}_{s}), \\
\textbf{P}_{r} =& \operatorname{LN}(\operatorname{MLP}(\textbf{P}^{i}) + \textbf{P}^{i}),   
\end{aligned}
\end{equation}
where $\textbf{P}^{i} \in \mathbb{R}^{N \times C}$ is the generated intermediate prototype, $\operatorname{LN}(\cdot )$ corresponds to the layer normalization, $\operatorname{MLP}(\cdot)$ represents the multilayer perception and $\operatorname{MHA}(\cdot)$ signifies the multi-head attention layer. We then conduct average-pooling and max-pooling over the representation $\textbf{P}_r$ in element dimension to preserve the most useful information which leads to new features $\mathcal{I}_{avg}^{inter}, \mathcal{I}_{max}^{inter} \in \mathbb{R}^{N \times 1}$. This can be simply expressed as:

%After that, we can aggregate core inter-prototype information maps $\mathcal{I}_{avg}^{inter}, \mathcal{I}_{max}^{inter} \in \mathbb{R}^{N \times 1}$ by performing average-pooling and max-pooling over the representation $\textbf{P}_r$ in element dimension, written as: 
\begin{equation}
\begin{cases}
\mathcal{I}_{avg}^{inter}= \operatorname{AvgPool}(\textbf{P}_{r})  \\
\mathcal{I}_{max}^{inter} = \operatorname{MaxPool}(\textbf{P}_{r}).
\end{cases}
\end{equation} 

The obtained features are concatenated along the channel dimension, enabling the multilayer perception block to be exploited to project the outcome into a single-channel feature map $\mathcal{I}^{inter}$, denoted as: 
\begin{equation}
    \mathcal{I}^{inter} = \operatorname{MLP}(\mathcal{I}_{avg}^{inter} \oplus \mathcal{I}_{max}^{inter}).
\end{equation} 
where $\mathcal{I}^{inter} \in \mathbb{R}^{N \times 1}$ will be activated by the sigmoid function and then multiplied by $\textbf{P}_{r}$ to produce the initial calibrated prototypical representation $\hat{\textbf{P}}_{\alpha}\in \mathbb{R}^{N \times C}$. Formally, it is expressed as: 
\begin{equation}
    \hat{\textbf{P}}_{\alpha} = \sigma (\mathcal{I}^{inter}) \odot  \textbf{P}_{r},
\end{equation}
where $\odot $ denotes the element-wise multiplication broadcast along the channel dimension, and $\sigma$ denotes the sigmoid activation function. The final output processed by the inter-prototype self-debiasing pathway is denoted as $\textbf{P}_{\alpha} = \operatorname{E}_{t}(\hat{\textbf{P}}_{\alpha}), \textbf{P}_{\alpha} \in \mathbb{R}^{N \times C}$, where $\operatorname{E}_{t}$ is the transformer encoder for enhancing the perceptive ability of $\hat{\textbf{P}}_{\alpha}$.   

%In order to further enhance perceptive ability of obtained $\hat{\textbf{P}}_{\alpha}$, the transformer encoder $\operatorname{E}_{t}$ is adopted to obtain final prototypical representation of inter-prototype self-debiasing manner: $\textbf{P}_{\alpha} = \operatorname{E}_{t}(\hat{\textbf{P}}_{\alpha}), \textbf{P}_{\alpha} \in \mathbb{R}^{N \times C}$.

The intra-prototype self-debiasing pathway is designed to run in parallel with the inter-prototype route and work in a similar manner. Formally,  
%Specifically, support prototypical representation $\tilde{\textbf{P}}_{s}$  is reconstructed by the encoder $\operatorname{E}_{t}$ to formulate $\textbf{P}_{r} = \operatorname{E}_{t}(\tilde{\textbf{P}}_{s})$. Intra-prototype information maps  $\mathcal{I}_{avg}^{intra}, \mathcal{I}_{max}^{intra} \in \mathbb{R}^{1 \times N}$ can be aggregated over $\textbf{P}_{r}$ with the average-pooling and max-pooling operations across the channel dimension. Then, information maps integration and activation process  similar to the operations in inter-prototype self-debasing manner are conducted,  obtaining the representation $\hat{\textbf{P}}_{\beta} \in \mathbb{R}^{N \times C}$ as:
\begin{equation}
    \begin{aligned}
        \mathcal{I}^{intra} &= \operatorname{MLP}(\mathcal{I}^{intra}_{avg} \oplus \mathcal{I}^{intra}_{max}), \\
        \hat{\textbf{P}}_{\beta} &= \sigma (\mathcal{I}^{intra}) \odot  \textbf{P}_{r},
    \end{aligned}
\end{equation}
where $\odot$ denotes the element-wise multiplication broadcast along the spatial dimension. The output of the intra-prototype self-debiasing is $\textbf{P}_{\beta} = \operatorname{E}_{t}(\hat{\textbf{P}}_{\beta}), \textbf{P}_{\beta} \in \mathbb{R}^{N \times C}$. 

\subsubsection{Interactive Debiasing}
\begin{figure}[!t]                   % htbp
\centering
\includegraphics[width=0.99\columnwidth]{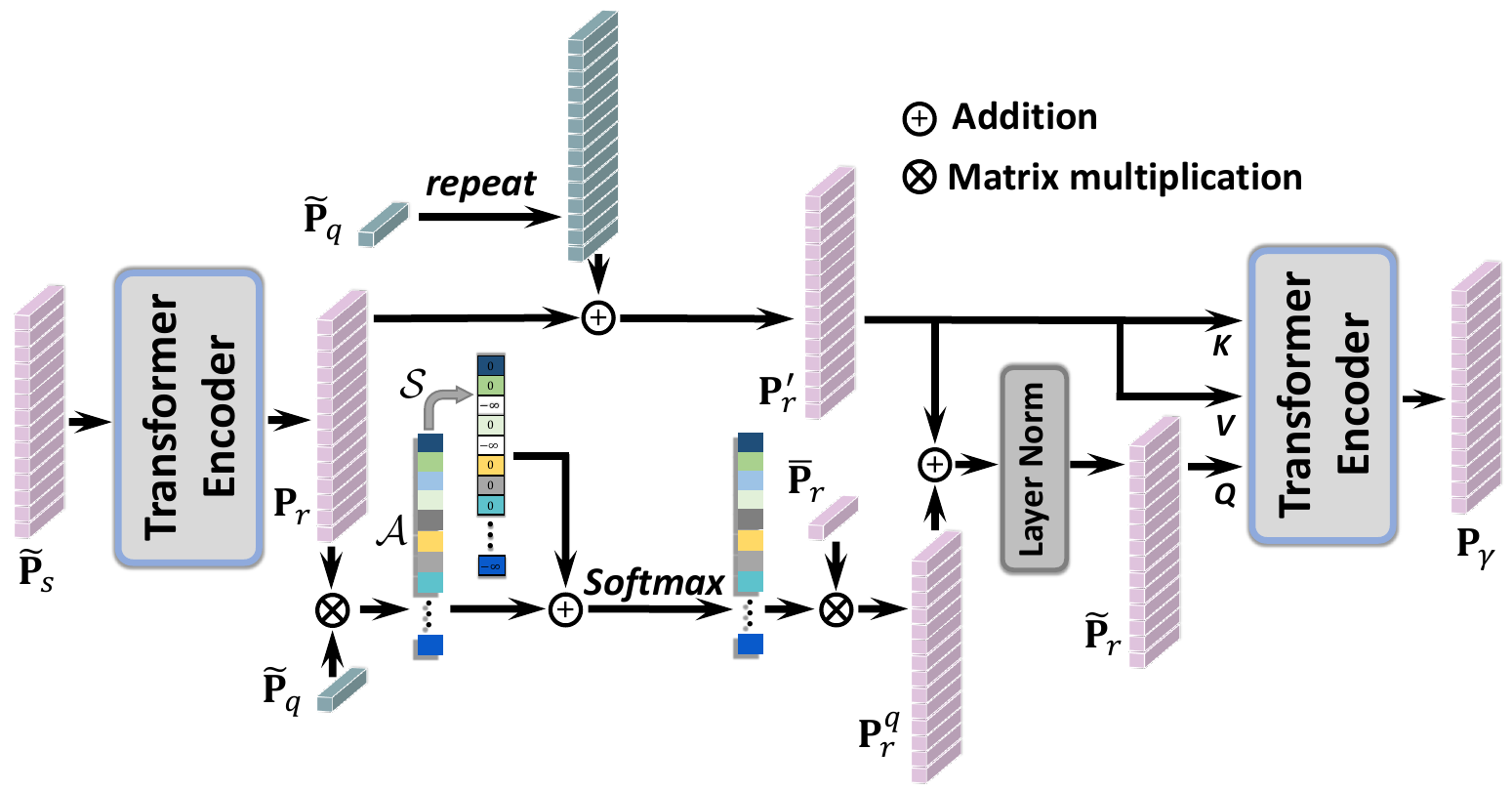}
\caption{Details of the Interactive Debiasing (ID) block. An affinity map $\mathcal{A}$ is calculated between the support prototypical representation and the query prototype. A self-selection mechanism is then applied to $\mathcal{A}$ to deliver the feature map $\mathcal{S}$ which is beneficial to suppress perturbing parts and reconstruct the representations. }
\label{ID block}
\end{figure}

We propose a novel Interactive Debiasing (ID) module which introduces the coarse query prototypes $\tilde{\textbf{P}}_{q} \in \mathbb{R}^{1 \times C}$ to interactively assist the support prototypes debiasing. 

As illustrated in Fig. \ref{ID block}, the prototypical representation $\tilde{\textbf{P}}_{s}$ (originated from $P_{s, enhanced}$) is first mapped into $\textbf{P}_{r} \in \mathbb{R}^{N \times C}$ by using the transformer encoder $\operatorname{E}_{t}$. Given the coarse prototype $\tilde{\textbf{P}}_{q}$ generated from the QPG module,  an affinity map $\mathcal{A} = \textbf{P}_{r}\tilde{\textbf{P}}_{q}^{\top}$,  where $\mathcal{A} \in \mathbb{R}^{N \times 1}$, is calculated to measure the correlations between the query and support representations. With the available affinity map $\mathcal{A}$, a self-selection mechanism is applied and results in the feature map $\mathcal{S} \in \mathbb{R}^{N \times 1}$. This can be written as:
\begin{equation}
\mathcal{S}_{i}(\mathcal{A}_{i}) = \begin{cases}
             0 & \text{ if } \mathcal{A}_{i} >= \xi  \\
 -\infty  & otherwise 
\end{cases},   i \in  \left \{  0,1, ... , N\right \},
\end{equation}
where $\xi$ is a threshold that can be determined by using $\xi = (min(\mathcal{A}) + mean(\mathcal{A}))/2$, $\mathcal{S}$ indicates the chosen regions from multiple regional prototypical representations that can be integrated with the query prototype. With the use of $\mathcal{S}$, the heterogeneous or perturbing regions of the support foreground will be suppressed by the \textit{Softmax} function to yield the prototypical representation $\textbf{P}_{r}^{q} \in  \mathbb{R}^{N \times C}$: 
\begin{equation}
\textbf{P}_{r}^{q} = \operatorname{softmax}(\mathcal{A} + \mathcal{S}) \bar{\textbf{P}}_{r},
\end{equation}
where $\bar{\textbf{P}}_{r}$ denotes the global prototype delivered by using the global pooling $\bar{\textbf{P}}_{r} = \operatorname{GlobalPool}(\textbf{P}_{r}), \bar{\textbf{P}}_{r} \in \mathbb{R}^{1 \times C}$. Then, $\textbf{P}_{r}^{q}$ will be added to $\textbf{P}_{r}$ together with $\tilde{\textbf{P}}_{q}$ for further query information interaction which will produce the representation $\tilde{\textbf{P}}_{r} \in \mathbb{R}^{N \times C}$: 
\begin{equation}
    \tilde{\textbf{P}}_{r}= \operatorname{LN}(\textbf{P}_{r}^{q} + \textbf{P}_{r}^{'}),
\end{equation}
where $\textbf{P}_{r}^{'} = \textbf{P}_{r} + \operatorname{repeat}(\tilde{\textbf{P}}_{q}), \textbf{P}_{r}^{'} \in $ and $\operatorname{repeat}(\cdot)$ denotes to repeat $\tilde{\textbf{P}}_{q}$ by N times to form a $(N, C)$ tensor. The interactive debiasing prototypical representation $\textbf{P}_{\gamma} \in \mathbb{R}^{N \times C}$ can be obtained by: 
\begin{equation}
    \begin{aligned}
       \textbf{P}^{ii} &= \operatorname{LN}(\operatorname{MHA}(\tilde{\textbf{P}}_{r}, \textbf{P}^{'}, \textbf{P}^{'}) + \tilde{\textbf{P}}_{r}), \\
        \textbf{P}_{\gamma} &= \operatorname{LN}(\operatorname{MLP}(\textbf{P}^{ii}) + \textbf{P}^{ii}).
    \end{aligned}
\end{equation}

\subsubsection{Prototype Regeneration}

\begin{figure}[!t]                   % htbp
\centering
\includegraphics[width=0.9\columnwidth]{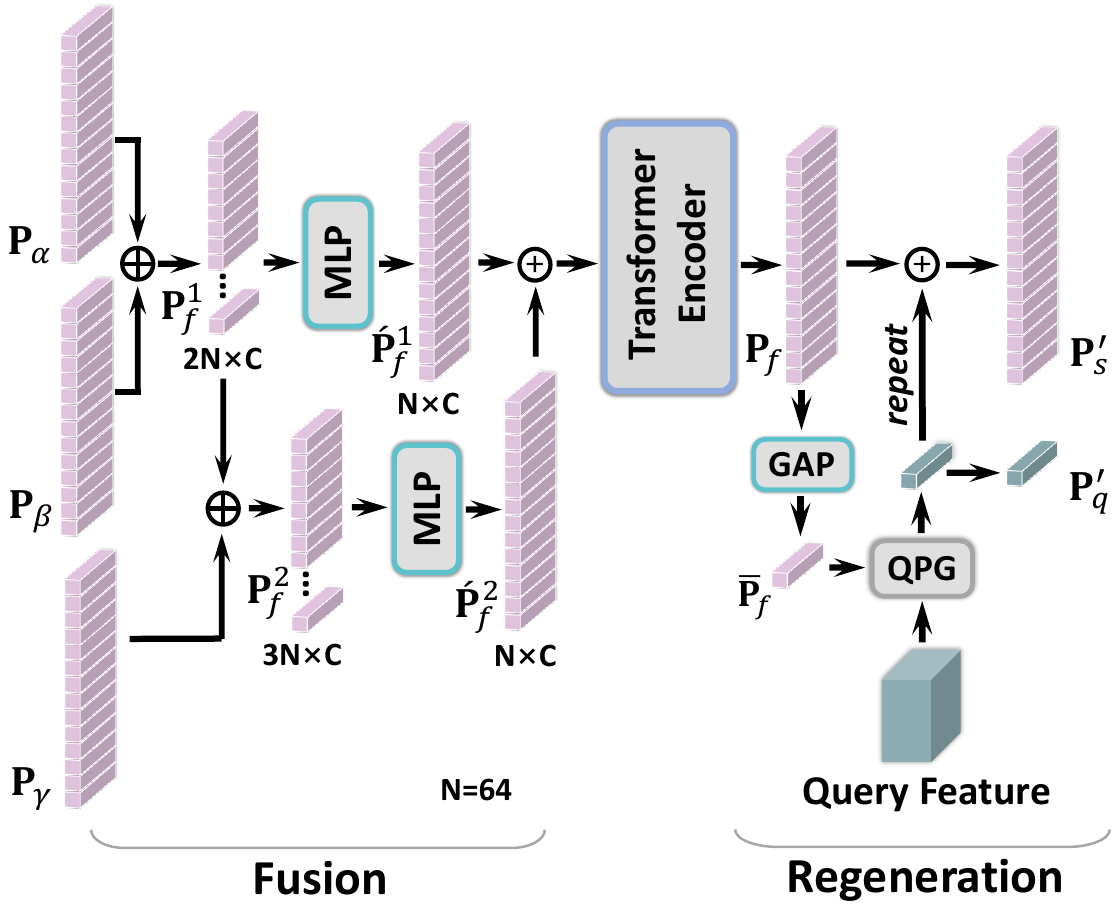}
\vspace{-1ex}
\caption{The Prototype Regeneration (PR) block consists of a \textit{Fusion} step and a \textit{Regeneration} step. Three types of prototypical representations $\textbf{P}_{\alpha}$, $\textbf{P}_{\beta}$ and $\textbf{P}_{\gamma}$ are first fused into an intermediate representation $\textbf{P}_f$ and then starts regeneration to produce the support prototypical representation $\textbf{P}_s^{'}$ and the query prototype $\textbf{P}_q^{'}$ used for the next PRD module.}
\vspace{-2ex}
\label{Re-Generation block}
\end{figure}

Given debiased prototypical representations $\textbf{P}_{\alpha}$, $\textbf{P}_{\beta}$ and $\textbf{P}_{\gamma}$, support prototypical representation $\textbf{P}_{s}^{'} \in \mathbb{R}^{N \times C}$ and coarse query prototype $\textbf{P}_{q}^{'} \in \mathbb{R}^{1 \times C}$ for the next PRD module are calculated by using the Prototype Regeneration (PR) module. As illustrated in Fig. \ref{Re-Generation block}, the PRG module has two main components: a fusion step and a regeneration step.

For the fusion step, we first concatenate the prototypical representations $\textbf{P}_{\alpha}, \textbf{P}_{\beta} \in \mathbb{R}^{N \times C}$ generated by the MS block to obtain the first fused representation $\textbf{P}_{f}^{1} \in \mathbb{R}^{2N \times C}$, which is calculated as follows:
\begin{equation}
    \textbf{P}_{f}^{1} = \textbf{P}_{\alpha} \oplus \textbf{P}_{\beta}.
\end{equation} 

Subsequently, the prototypical representation $\textbf{P}_{\gamma} \in \mathbb{R}^{N \times C}$ generated from the ID block is concatenated with $\textbf{P}_{f}^{1}$ to get the second fused representation $\textbf{P}_{f}^{2} \in \mathbb{R}^{3N \times C}$, written as: 
\begin{equation}
    \textbf{P}_{f}^{2} = \textbf{P}_{f}^{1} \oplus \textbf{P}_{\gamma}.
\end{equation}

The coarse fused prototypical representation $\textbf{P}_{f,coarse} \in \mathbb{R}^{N \times C}$ simply aggregates the projected $\operatorname{MLP}(\textbf{P}_{f}^{1})$ and $\operatorname{MLP}(\textbf{P}_{f}^{2})$ where $\acute{\textbf{P}}_{f}^{1}$ and $\acute{\textbf{P}}_{f}^{2}$ are the mapped intermediate fused prototypical representations. Formally,  
%then both $\textbf{P}_{f}^{1}$ and $\textbf{P}_{f}^{2}$ are mapped into the intermediate fused prototypical representations $\acute{\textbf{P}}_{f}^{1}, \acute{\textbf{P}}_{f}^{2} \in \mathbb{R}^{N \times C}$ using the multilayer perceptron block and added together to obtain the coarse fused prototypical representation $\textbf{P}_{f,coarse} \in \mathbb{R}^{N \times C}$:
\begin{equation}
\begin{aligned}
    \textbf{P}_{f, coarse} &= \acute{\textbf{P}}_{f}^{1} + \acute{\textbf{P}}_{f}^{2} \\
    &= \operatorname{MLP}(\textbf{P}_{f}^{1}) +  \operatorname{MLP}(\textbf{P}_{f}^{2}). 
\end{aligned}
\end{equation} 
where $\textbf{P}_{f}$ is used to produce the final fused prototypical representation $\textbf{P}_{f}$ by using the vanilla transformer encoder $\operatorname{E}_{t}$, represented as:
%Consequently, the final fused prototypical representation $\textbf{P}_{f}$ is obtained with the vanilla transformer encoder $\operatorname{E}_{t}$:
\begin{equation}
    \textbf{P}_{f} = \operatorname{E}_{t}(\textbf{P}_{f,coarse}), \textbf{P}_{f} \in \mathbb{R}^{N \times C}.
    \end{equation}

The regeneration step aims to produce the query prototype $\textbf{P}_{q}^{'}$ with the QPG module introduced in Section \ref{sec3c}, which can be computed by:
\begin{equation}
    \textbf{P}_{q}^{'} = \operatorname{QPG}(\textbf{F}_{q}, \bar{\textbf{P}}_{f}),
\end{equation}
where $\bar{\textbf{P}}_{f} = \operatorname{GAP}(\textbf{P}_{f}), \bar{\textbf{P}}_{f} \in \mathbb{R}^{N \times C}$ is received by using the global average pooling operation over $\textbf{P}_{f}$, and the support prototypical representation for the next PRD module can be obtained as:
\begin{equation}
    \textbf{P}_{s}^{'} = \textbf{P}_{f} + \operatorname{repeat}(\textbf{P}_{q}^{'}).
\end{equation}

\subsection{Assembled Prediction}
The debiased support prototypical representation $\textbf{P}_{s}$ and the regenerated query prototype $\textbf{P}_{q}$ can be achieved by processing the $\tilde{\textbf{P}}_{s}$ and $\tilde{\textbf{P}}_{q}$ with $M$ stacked PRD modules. Formally, this can be denoted as:
\begin{equation}
    \left \{ \textbf{P}_s, \textbf{P}_q \right \}  = \operatorname{PRD}^{M}(\tilde{\textbf{P}}_{s}, \tilde{\textbf{P}}_{q}).
\end{equation}

We can then infer the predictions for the query by: 
\begin{equation}
    \begin{cases}
\mathcal{M}_s^{f} = 1 - \sigma (S(\textbf{F}_{s}, \operatorname{GlobalPool}(\textbf{P}_s)) - \tau);  \\
\mathcal{M}_q^{f} = 1 - \sigma (S(\textbf{F}_{q}, \textbf{P}_q) - \tau),  
\end{cases}
\end{equation} 
and the final predicted foreground can be obtained by:
\begin{equation}\label{AP}
    \mathcal{M}^f = \lambda  \mathcal{M}_s^f + (1 - \lambda)\mathcal{M}_q^f,
\end{equation} 
where $\lambda$ is the assembling coefficient, and is set to 0.7 (more details can be found in the experiment section).  

%\subsection{Objective Loss}
The binary cross-entropy loss $\mathcal{L}_{ce}$ is adopted to determine the error between the predict masks $(\mathcal{M}^{f}, \mathcal{M}^{b})$ and the given ground-truth $(\tilde{\mathcal{M}}^{f}, \tilde{\mathcal{M}}^{b})$. Formally, 
\begin{equation}
\mathcal{L}_{ce} = -\frac{1}{HW}\sum_{i,j} \tilde{\mathcal{M}}^{f}_{(i,j)}log(\mathcal{M}^{f}_{(i,j)}) + \tilde{\mathcal{M}}^{b}_{(i,j)}log(\mathcal{M}^{b}_{(i,j)}).
\end{equation}

\section{Experiments}\label{sec:exp}

\subsection{Experimental Setting}

\subsubsection{Datasets}
We comprehensively evaluate the proposed method on three publicly available datasets with different modalities and  anatomical structures, including (a) \textbf{CHAOS}: an abdominal MRI dataset published in ISBI 2019 Combined Healthy Abdominal Organ Segmentation Challenge \cite{kavur2021chaos}, (b) \textbf{SABS}: an abdominal CT dataset from MICCAI 2015 Multi-Atlas Abdomen Labeling Challenge \cite{landman2015ct} and (c) \textbf{CMR}: a cardiac MRI dataset from MICCAI 2019 Multi-Sequence Cardiac MRI Segmentation Challenge \cite{zhuang2018cmr}. \textbf{CHAOS} and \textbf{SABS} share the same categories of labels which are \textit{liver}, \textit{spleen}, \textit{left kidney} (LK) and \textit{right kidney} (RK). \textbf{CMR} dataset has three categories of labels including \textit{left ventricular myocardium} (LV-MYO), \textit{right ventricle blood pool} (LV-BP) and \textit{right ventricle} (RV). Considering the efficiency of method training, we here reformat the 3D scans of these datasets into 2D axial and 2D short-axis slices with size of $256 \times 256$ to segment 3D images using a 2D method.

\subsubsection{Implementation Details}
Before method training, the pseudo masks for training scans are generated using the super-voxel clustering method according to \cite{hansen2022anomaly}. To conduct episodic training for meta-learning method,  we select the support slices and query slices based on the strategy in \cite{roy2020squeeze}. For each 3D image in both query and support sets, we systematically subdivide the region-of-interest into 3 equally-sized segments. Each query segment's corresponding support samples consist of central slices from matching segments across all support scans. The five folds cross validation is conducted following the protocol in \cite{hansen2022anomaly}. 

Our method is implemented using PyTorch (v1.10.2) based on the SSL-ALPNet \cite{ouyang2022self} and ADNet \cite{hansen2022anomaly} implementations. The ResNet101 \cite{he2016deep} pretrained on part of MS-COCO dataset is employed as the backbone of feature extractor $f_{\theta}$. Data pre-processing pipeline is based on \cite{ouyang2022self,hansen2022anomaly}. Each single channel 2D slice will be reproduced by three times and concatenated along channel for suitable for convolutional network input. The method is trained using 1-way 1-shot configuration for over 50K iterations  with the SGD optimizer. During training phase, the initial learning rate is set to $1 \times 10^{-3}$ with a step decay of 0.8 each 1000 iteration.

\subsubsection{Evaluation Protocol}
For evaluation purposes, Dice Similarity Coefficient (DSC) score is employed to compare the predictions given by method with the ground truth of segmentations. Furthermore, we employ two settings to challenge the proposed method in terms of ability of generalizing into new data. Specifically, two settings are: \textbf{Setting 1}: some of slices containing the test classes might appear during the training phase; \textbf{Setting 2}: those slices containing test classes are completely removed during training phase. It is worth noticing that Setting 2 cannot be implemented on scans from CMR dataset, theoretically because all organ classes are likely to be present on one slice simultaneously.

\subsection{Comparison with State-of-the-Arts}

\subsubsection{Quantitative Results}

\begin{table*}[!t]
\centering
\caption{Quantitative Comparison DSC score ($\%$) of different methods under \textbf{Setting1} and \textbf{Setting2} on CHAOS and SABS. The best value is shown in bold font, and the second-best value is underlined. '--' means not reported.}
\label{tab1}
\resizebox{\textwidth}{!}{%
\begin{tabular}{c|l|l|ccccc|ccccc}
\toprule[1pt]
\midrule
\multirow{2}{*}{Setting} & \multirow{2}{*}{Method} & \multicolumn{1}{c|}{\multirow{2}{*}{Ref.}} & \multicolumn{5}{c|}{CHAOS} & \multicolumn{5}{c}{SABS} \\ \cmidrule{4-13} 
 &  & \multicolumn{1}{c|}{} & LK & RK & Spleen & Liver & Mean & LK & RK & Spleen & Liver & Mean \\ \midrule
\multirow{9}{*}{\textbf{1}} & PANet\cite{wang2019panet} & CVPR'19 & 30.99 & 32.19 & 40.58 & 50.40 & 38.53 & 20.67 & 21.19 & 36.04 & 49.55 & 32.86 \\
 & SENet \cite{roy2020squeeze} & MIA'20 & 45.78 & 47.96 & 47.30 & 29.02 & 42.51 & 24.42 & 12.51 & 43.66 & 35.42 & 29.00 \\
 & SSL-ALPNet \cite{ouyang2022self} & ECCV'20 & 81.92 & 85.18 & 72.18 & 76.10 & 78.84 & 72.36 & 71.81 & 70.96 & 78.29 & 73.35 \\
 & ADNet \cite{hansen2022anomaly} & MIA'22 & 73.86 & 85.80 & 72.29 & \underline{82.11} & 78.51 & 72.13 & \underline{79.06} & 63.48 & 77.24 & \underline{77.24} \\
 & AAS-DCL \cite{wu2022dual} & ECCV'22 & 80.37 & 86.11 & \underline{76.24} & 72.33 & 78.76 & 74.58 & 73.19 & 72.30 & \underline{78.04} & 74.52 \\
 & SR\&CL \cite{wang2022few} & MICCAI'22 & 79.34 & 87.42 & 76.01 & 80.23 & 80.77 & 73.45 & 71.22 & \textbf{73.41} & 76.06 & 73.53 \\
 & CRAPNet \cite{ding2023few} & WACV'23 & \textbf{81.95} & 86.42 & 74.32 & 76.46 & 79.79 & \underline{74.69} & 74.18 & 70.37 & 75.41 & 73.66 \\
 &Q-Net \cite{shen2023qnet} & IntelliSys'23         &78.36       & \underline{87.98}         &75.99         &  81.74    & \underline{81.02}           & --        &--         & --         & --      & --                  \\
 \rowcolor{gray!25}&  PAMI (Ours)  &  &\underline{81.83}  &\textbf{88.73}  &\textbf{76.37} & \textbf{82.59} & \textbf{82.38}  & \textbf{76.52}  & \textbf{80.57}  & \underline{72.38}  & \textbf{81.32}& \textbf{77.69} \\ \midrule
\multirow{9}{*}{\textbf{2}} & PANet\cite{wang2019panet} & CVPR'19 & 53.45 & 38.64 & 50.90 & 42.26 & 46.33 & 32.34 & 17.37 & 29.59 & 38.42 & 29.43 \\
 & SENet \cite{roy2020squeeze} & MIA'20 & 62.11 & 61.32 & 51.80 & 27.43 & 50.66 & 32.83 & 14.34 & 0.23 & 0.27 & 11.91 \\
 & SSL-ALPNet \cite{ouyang2022self} & ECCV'20 & 73.63 & 78.39 & 67.02 & 73.05 & 73.02 & 63.34 & 54.82 & 60.25 & \underline{73.65} & 63.02 \\
 & ADNet \cite{hansen2022anomaly} & MIA'22 & 59.64 & 56.68 & 59.44 & 77.03 & 63.20 & 48.41 & 40.52 & 50.97 & 70.63 & 52.63 \\
 & AAS-DCL \cite{wu2022dual} & ECCV'22 & \underline{76.90} & 83.75 & \underline{74.84} & 69.94 & 76.36 & 64.71 & \textbf{69.95} & 66.36 & 71.61 & 68.16 \\
 & SR\&CL \cite{wang2022few} & MICCAI'22 & \textbf{77.07} & \underline{84.24} & 73.73 & 75.55 & \underline{77.65} & 67.39 & 63.37 & 67.36 & 73.63 & 67.94 \\
 & CRAPNet \cite{ding2023few} & WACV'23 & 74.66 & 82.77 & 70.82 & 73.82 & 75.52 & \underline{70.91} & 67.33 & \underline{70.17} & 70.45 & \underline{69.72}\\
 &Q-Net \cite{shen2023qnet} & IntelliSys'23         &64.81       &65.94         &65.37         &  \underline{78.25}    & 68.59            &--       &   --     &--        &--        & --            \\
 \rowcolor{gray!25} & PAMI (Ours) &  &  74.51 & \textbf{86.73}  & \textbf{75.80}  & \textbf{81.09} & \textbf{79.53} & \textbf{72.36} & \underline{67.54}  & \textbf{71.95}  & \textbf{74.13}  & \textbf{71.49}  \\ \midrule
 \bottomrule[1pt]
\end{tabular}%
}
\end{table*}

\begin{figure*}[ht]                   % htbp
\centering
\includegraphics[width=1\textwidth]{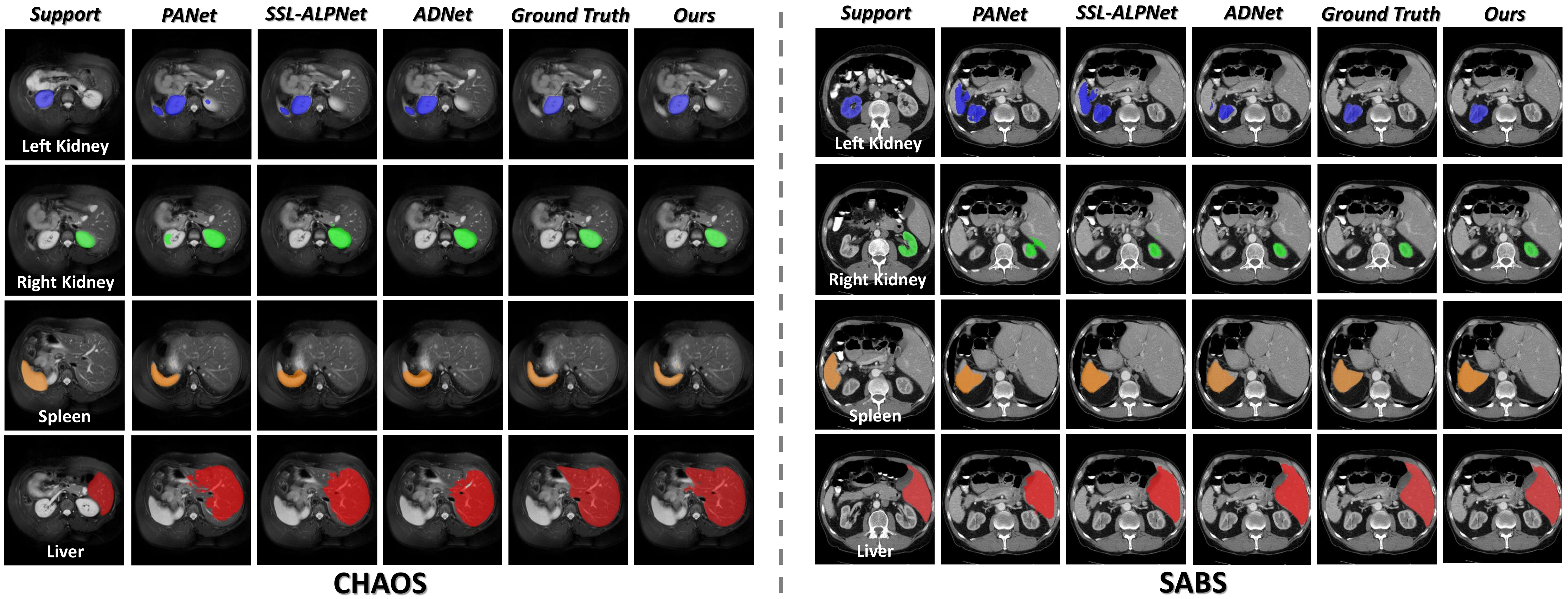}
\vspace{-4ex}
\caption{The qualitative results of our method and other baseline methods on the \textbf{CHAOS} dataset and the \textbf{SABS} dataset. }
\vspace{-2ex}
\label{results in CHAOS and SABS}
\end{figure*}

As shown in Table \ref{tab1} and Table \ref{tab2}, we compare the performance of our proposed method with some baseline methods including the vanilla PANet \cite{wang2019panet}, SENet \cite{roy2020squeeze}, SSL-ALPNet \cite{ouyang2022self}, ADNet \cite{hansen2022anomaly}, AAS-DCL \cite{wu2022dual}, SR\&CL \cite{wang2022few}, CRAPNet \cite{ding2023few} and Q-Net \cite{shen2023qnet} under two experimental settings on three datasets \textbf{CHAOS}, \textbf{SABS} and \textbf{CMR}. In Table \ref{tab1}, we illustrate the mean DSC score of five cross-validation folds on four organs (LK, RK, Spleen and Liver) and the mean DSC score of these four organs results under two experimental settings (setting 1\&2). Our proposed method first outperforms all of other methods in mean DSC score of four organs under these two experimental settings. Specifically, the proposed method achieve the highest mean DSC scores of 82.38\% and 79.53\% on the CHAOS dataset under setting 1 and setting 2, surpassing the second-best methods (Q-Net and SR\&CL) by 1.36\% and 1.88\% respectively. On the SABS dataset, the proposed method also attains the optimal DSC scores on organs including right kidney (RK), left kidney (LK) and spleen. The mean DSC score on the SABS dataset outperforms the second-best result by 0.45\% and 1.77\% under setting 1 and setting 2, respectively. Additionally, testing on the CMR dataset, our method also exhibits a remarkable improvements on the three regions (LV-BP, LV-MYO and RV). As shown in Table \ref{tab2}, the proposed method obtains the optimal DSC scores of 89.57\%, 66.82\%, 80.17\% on LV-BP, LV-MYO, RV, achieving improvements of 0.08\%, 0.21\% on LV-MYO and RV, respectively, and only has 0.68\% degradation on LV-BP compared to Q-Net. However, we can obtain 0.7\% improvement compared on the metric of the mean value.

\begin{table}[!ht]
		\centering
		\caption{Quantitative Comparison of different methods DSC score (\%)  on CMR dataset under \textbf{Setting 1}. The best value is shown in bold font, and the second best value is underlined.}
        \label{tab2}
        \resizebox{0.49\textwidth}{!}{
        \begin{tabular}{l|l|cccc}
        \toprule[1pt]
        \midrule
        \multicolumn{1}{c|}{Method} &\multicolumn{1}{c|}{Ref.} & LV-BP & LV-MYO & RV    & Mean  \\ \midrule
       SENet \cite{roy2020squeeze} & CVPR'19             & 58.04 & 25.18  & 12.86 & 32.02 \\
       PANet \cite{wang2019panet} & MIA'20              & 72.77 & 44.76  & 57.13 & 58.20 \\
        SSL-ALPNet \cite{ouyang2022self} & ECCV'20         & 83.99 & \underline{66.74} & \underline{79.96} & 76.90 \\
        ADNet \cite{hansen2022anomaly} & MIA'22               & 87.53 & 62.43  & 77.31 & 75.76 \\
        AAS-DCL \cite{wu2022dual} & ECCV'22            & 85.21 & 64.03  & 79.13 & 76.12 \\
        SR\&CL \cite{wang2022few} & MICCAI'22           & 84.74 & 65.83  & 78.41 & 76.32 \\
        CRAPNet \cite{ding2023few} & WACV'23            & 83.02 & 65.48  & 78.27 & 75.59 \\
        Q-Net \cite{shen2023qnet} & IntelliSys'23         &\textbf{90.25}        &65.92        & 78.19      &\underline{78.15}         \\
       \rowcolor{gray!25}  PAMI (Ours)   &      & \underline{89.57}  & \textbf{66.82}   & \textbf{80.17}  &\textbf{78.85}  \\ \midrule
        \bottomrule[1pt]
        \end{tabular}}	
        \vspace{-5ex}
\end{table}

\begin{figure}[!h]                   % htbp
\centering
\includegraphics[width=1\columnwidth]{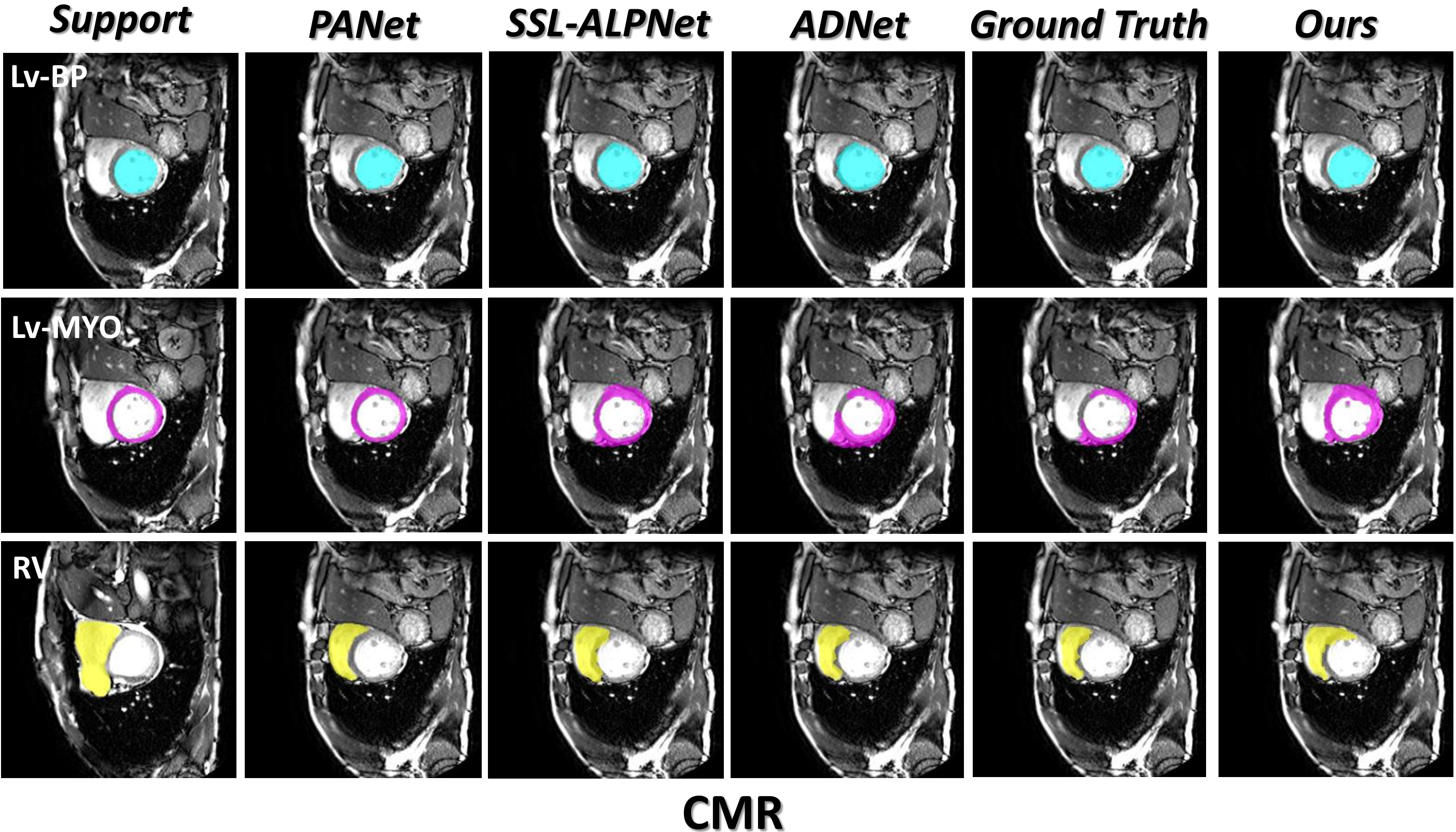}
\caption{The qualitative results of our method and other baseline methods on the \textbf{CMR} dataset. }
\label{results in CMR}
\end{figure}

\subsubsection{Qualitative Results}

To analysis segmentation performance of our method intuitively, we present the qualitative results of our method and other baseline methods on three datasets CHAOS, SABS and CMR in Fig. \ref{results in CHAOS and SABS} and Fig. \ref{results in CMR}. We compare our segmentation results with the methods PANet, SSL-ALPNet and ADNet, which have provided the complete code repositories. On the CHAOS dataset, segmentation results of our method have the better border accuracy on the left kidney and spleen organs. Especially, only our method can segment the entire region boundary on the liver organ against other segmentation methods. On the left kidney and right kidney segmentation, our method can produce a finer segmentation performance in the edge details. On the SABS dataset, the proposed method significantly optimizes the segmentation outcomes of both the spleen and the liver, providing more refined organ boundaries and clearer textures compared to other baseline methods. Overall, these qualitative results provide the evidences for the effectiveness of the proposed method,  and highlights its potential as a valuable tool in clinical diagnosis and treatment under the limited data scenario.

\subsection{Ablation Study}

In this section, we conduct the ablation study for analysing the effectiveness of each components and settings of hyper-parameter in our method. All of ablation experiments are implemented on the CHAOS dataset under the setting 2. 

\begin{figure}[!t]                   % htbp
\centering
\includegraphics[width=0.9\columnwidth]{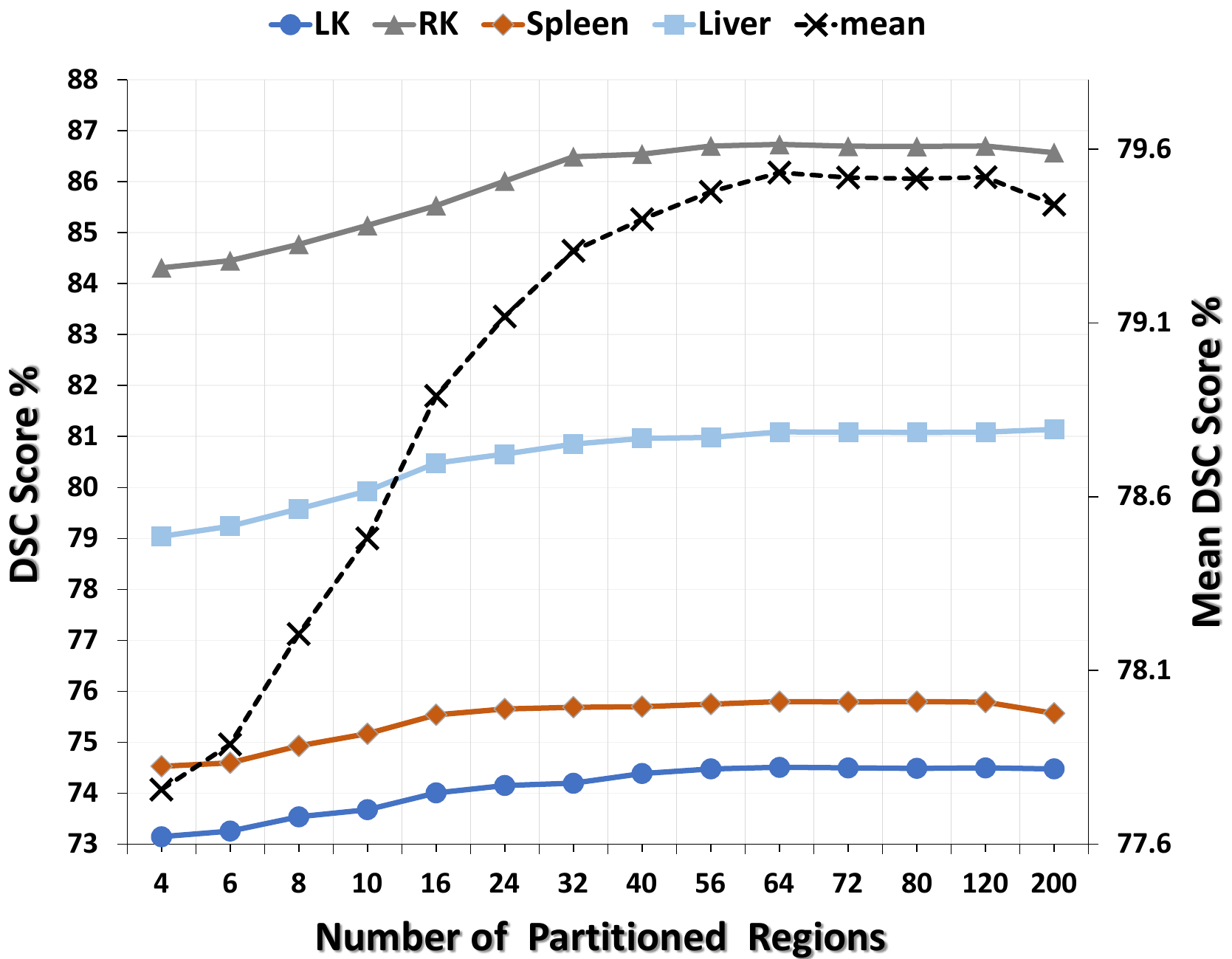}
\caption{Analysis of partitioned regions settings $N_f$. The DSC score of each organs (LK, RK, Spleen and Liver) and the mean DSC score of them under a series of $N_f$ settings are illustrated. }
\label{Ablation of N_f}
\end{figure}

\subsubsection{Analysis of Partitioned Regions $N_f$}

We analysis the performance of proposed method with the different settings of partitioned regions number $N_f$ in Fig. \ref{Ablation of N_f}. We evaluate the each DSC scores results on four organs (LK, RK, spleen and liver) and mean DSC score of them under a series of $N_f$ settings. As shown in Fig. \ref{Ablation of N_f}, each DSC scores of LK, RK, spleen and liver increase concomitantly with the increase of partitioned regions number $N_f$. In details, the DSC scores have a rapid growth from $N_f=4$ to $N_f = 56$, and it tends to flatten out and achieves highest around setting of $N_f = 64$. Then the DSC score will not further show growth with the increase of $N_f$ value, even the DSC score of spleen organ shows a noticeable decline when $N_f$ continues to increase to value of 200. This phenomenon reveals that more partitions can produce more representative sub-regions, thus purifying the learned prototype by eliminating those perturbing sub-regions. However, too many subregions imply excessive segmentation, which may introduce segmentation noise and lead to performance degradation.
%Based on this observation, the value of $N_f$ is set as 64 in our method for optimal and stable performance. 

\begin{figure}[!t]                   % htbp
\centering
\includegraphics[width=0.9\columnwidth]{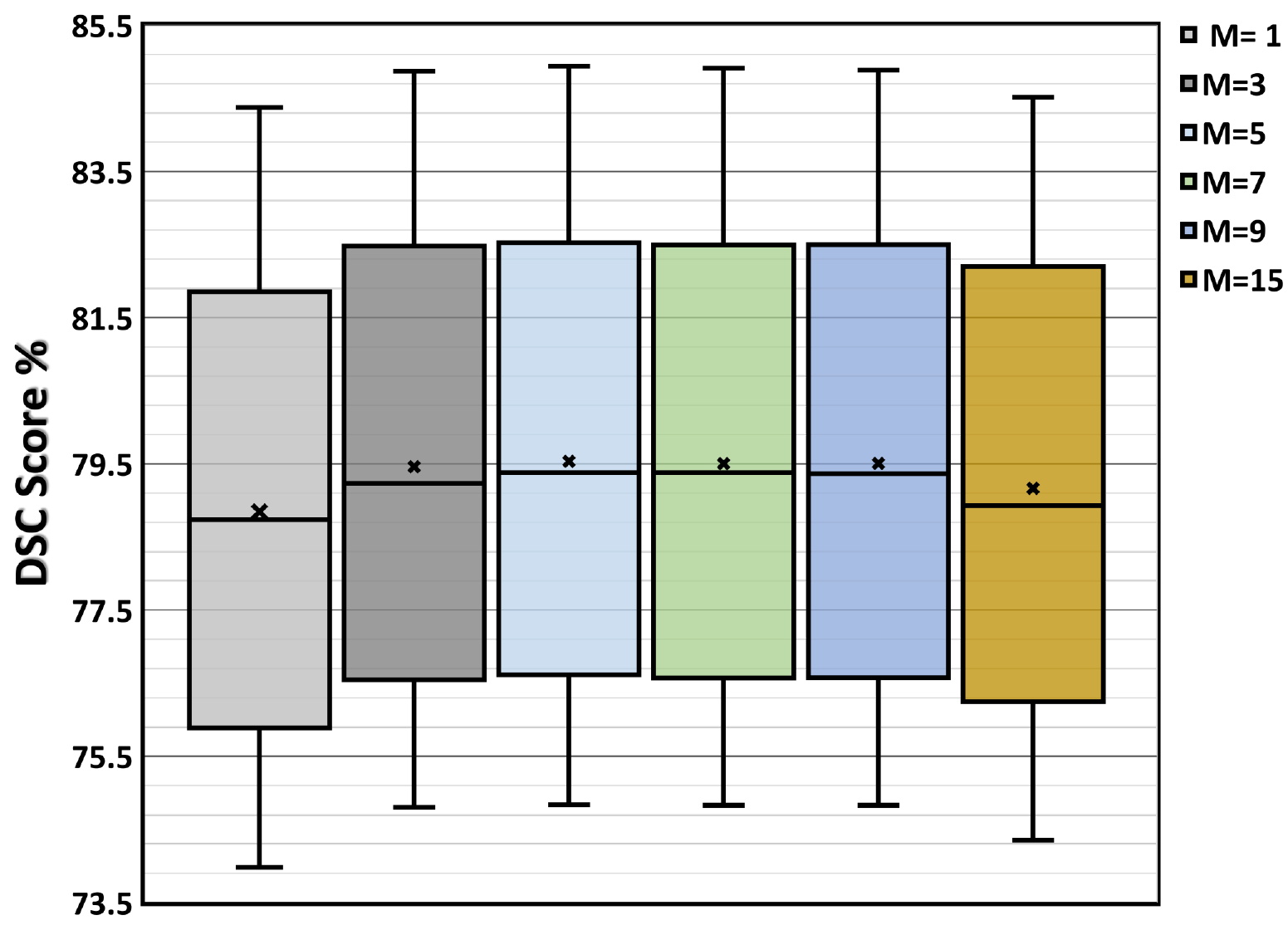}
\vspace{-1ex}
\caption{The box-plot for mean DSC scores of five cross-validation folds under different settings of $M (M=\{1,3,5,7,9,15\})$. }
\vspace{-2ex}
\label{Box plot of M}
\end{figure}

\begin{table}[!t]
\centering
		\caption{Details of mean DSC scores (\%) of each cross-validation folds under different values of $M$.}
        \label{tab3}
\begin{tabular}{c|c|c|c|c|c|c}
\toprule[1pt]
\midrule
\textbf{M}  & fold-0 & fold-1 & fold-2 & fold-3 & fold-4 & Mean   \\ \midrule
\textbf{1}        & 79.337 & 73.983 & 84.372 & 78.741 & 77.791 & 78.845 \\ \midrule
\textbf{3}          & 80.087 & 74.804 & 84.873 & 79.231 & 78.295 & 79.458 \\ \midrule
\rowcolor{gray!25}\textbf{5}          & 80.106 & \textbf{74.838} & \textbf{84.941} & 79.376 & \textbf{78.399} & \textbf{79.532} \\ \midrule
\textbf{7}          & 80.072& 74.829 & 84.912 & \textbf{79.378} & 78.316 & 79.501 \\ \midrule
\textbf{9}         & \textbf{80.113} & 74.831 & 84.883 & 79.365 & 78.326 & 79.503 \\ \midrule
\textbf{15}         & 79.892 & 74.354 & 84.513 & 78.929 & 78.142 & 79.166 \\ \midrule
\bottomrule[1pt]
\end{tabular}
\end{table}

\subsubsection{Effect of Prototypical Representation Debiasing Module}

\begin{figure}[!t]                   % htbp
\centering
\includegraphics[width=0.83\columnwidth]{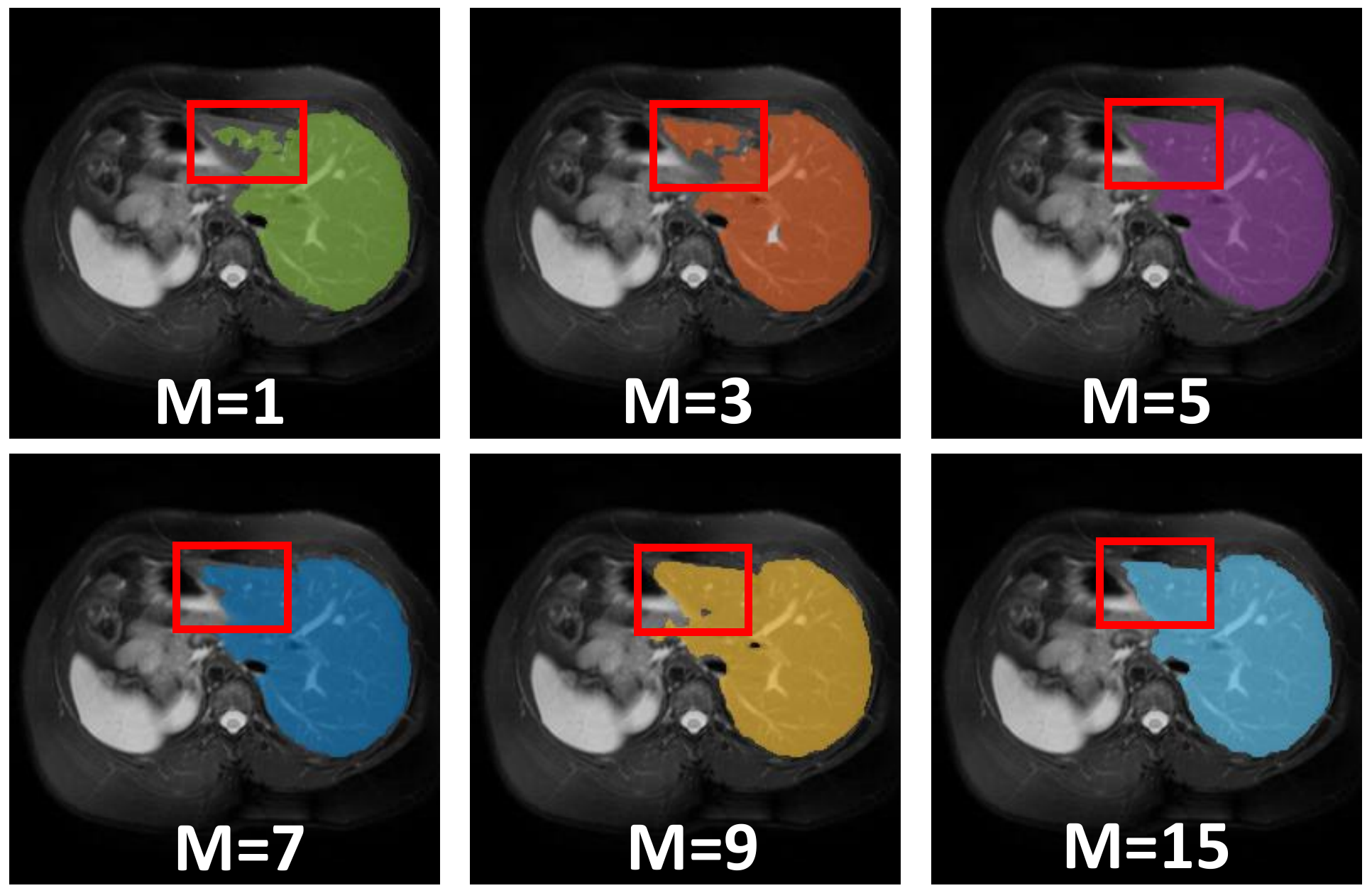}
\caption{The results of liver segmentation on the \textbf{CHAOS} dataset under the setting of $M=\{1, 3, 5, 7, 9, 15\}$.}
\label{effect of M}
\end{figure}

In our proposed method, the core debiasing operation employs a stacked of $M$ PRD modules. For conducting effectiveness analysis, we adopt the mean DSC scores of each cross-validation fold as the evaluation protocol, and the results of different settings of $M$ are presented in Fig. \ref{Box plot of M} and Table \ref{tab3}. By observing the box plot, we can analyse the distribution and variability of DSC scores of each folds under the different settings of $M$. Specifically, from $M=1$ to $M=5$, the DSC scores show a continuous increase in median and mean, which shows the effect of PRD module in improving performance. The distribution concentration of data is gradually improved and DSC score median of five cross-validation folds increases from around 78.8\% to around 79.48\%. After that, with the increase of number of PRD modules (from $M=7$ to $M=15$), the mean DSC score of five cross-validation folds fluctuates around 79.50\%, and it has no obvious signs of growth. Additionally, DSC score shows a decrease of almost 0.36\% by setting $M=15$, which means that score may have converged around setting $M=5$ and more stacked PRD modules lead to the overfitting. The box plot shows that, from $M=5$ to $M=9$, the data distribution concentration abilities of these settings have a little difference and medians are all close to the mean DSC scores. Moreover, the setting of $M=5$ gains the highest DSC score in mean (79.532\%), fold-1 (74.838\%), fold-2 (84.941\%), fold-3 (79.376\%) and fold-4 (78.399\%), and the best distribution concentration performance. Additionally, the qualitative results of liver segmentation under different $M$ settings are also shown in Fig. \ref{effect of M} for more intuitive comparison.

\subsubsection{Analysis of Assembling Coefficient $\lambda$}

\begin{figure}[!t]                   % htbp
\centering
\includegraphics[width=0.8\columnwidth]{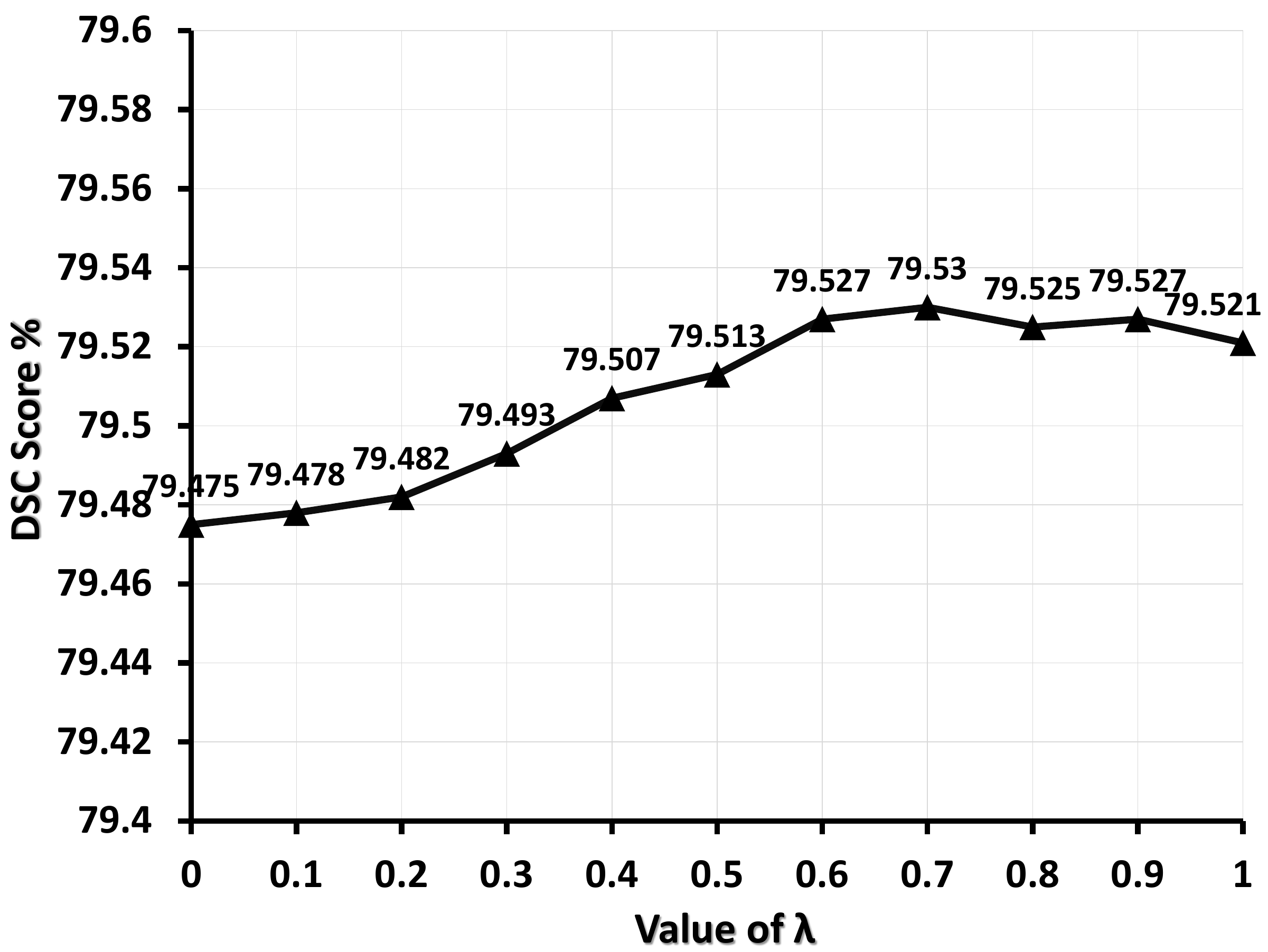}
\vspace{-1ex}
\caption{The curve of mean DSC score of four segmentation organs under different settings of coefficient $\lambda$.}
\label{curve of lambda}
\end{figure}

\begin{figure}                 % htbp
\centering
\includegraphics[width=0.83\columnwidth]{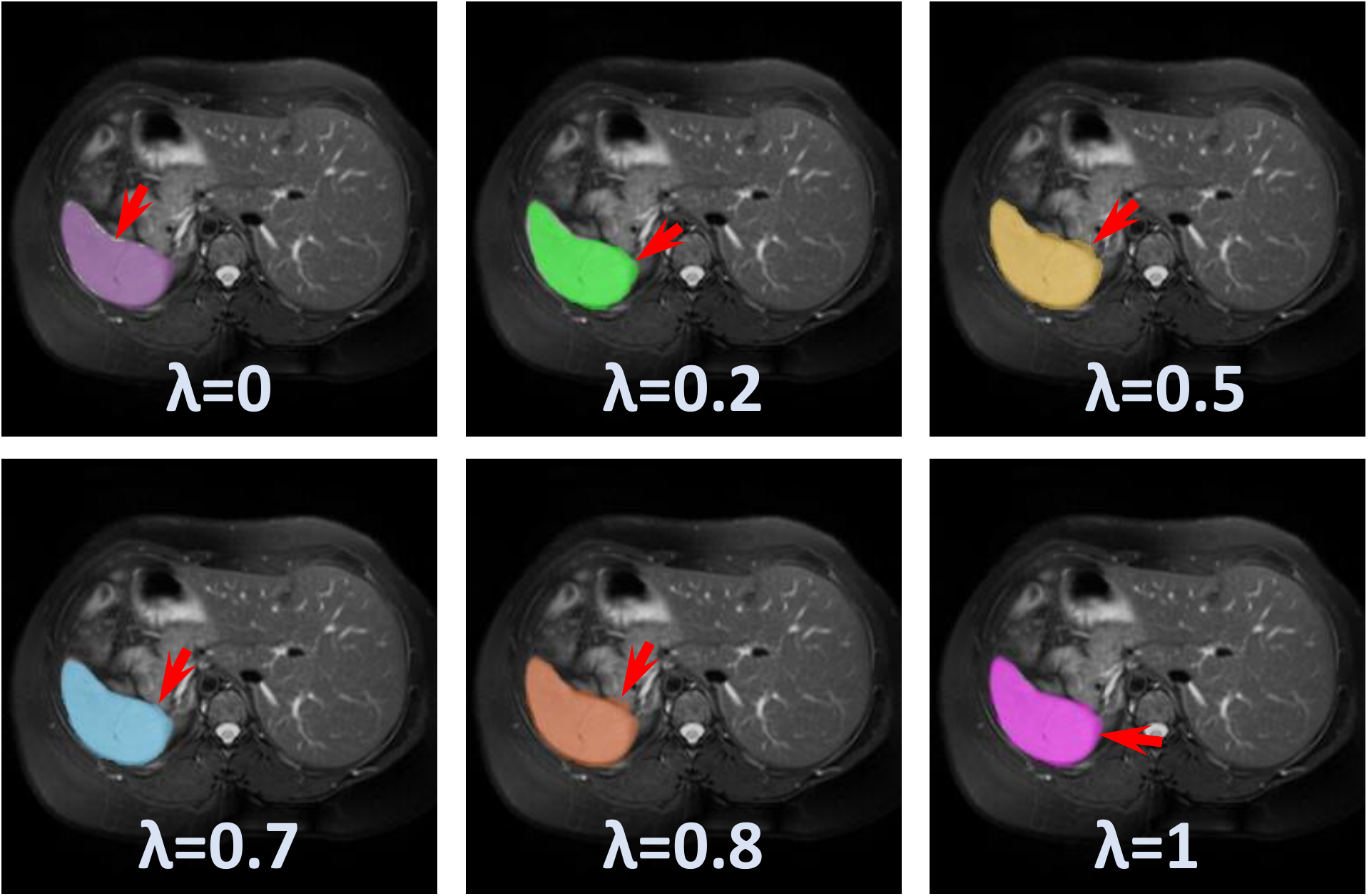}
\caption{The results of spleen segmentation in \textbf{CHAOS} dataset under the settings of $\lambda = \{0, 0.2, 0.5, 0.7, 0.8, 1\}$. }
\label{illustration of lambda}
\end{figure}

In this section, we discuss the effectiveness of different settings of assembling coefficient $\lambda$ used in the AP module. As illustrated in Fig. \ref{curve of lambda}, the mean DSC score of four segmentation organs on the CHAOS dataset shows a significant growth with the increase of value of $\lambda$, which means that the prediction map $\mathcal{M}_q^{f}$ using single query prototype information is not the optimal choice and more information from support prototypical representations should improve the accuracy. Based on this tendency, we can obtain the mean DSC score over 79.5\% after setting $\lambda = 0.5$ and the best mean DSC score of 79.53\% can be achieved around setting of $\lambda = 0.7$. As the value of $\lambda$ continues to increase, the mean DSC score does not show the corresponding growth although more support prototypical information is included. 
%Thus, we choose the $\lambda = 0.7$ as final setting in AP module. 
In addition, we also give the segmentation results in spleen organ under the setting of $\lambda = \{0, 0.2, 0.5, 0.7, 0.8, 1\}$ in Fig. \ref{illustration of lambda} which shows the details of segmentation region under different settings.

\section{Conclusion And Discussion}

In this work, we focus on the hard intra-class variation problem in few-shot medical image segmentation task. Here we introduce to partition the image into multiple sub-regions, and suppress the perturbing sub-regions of foreground of support image in prototypical level and then refine the rest part information into ideal prototypical representations for fast adapting. 
%Concretely, we design the novel Regional Prototype Learning (RPL) module to first partition the foreground of support into multiple sub-regions with Voronoi-based method and obtain multiple regional support prototypical representations. Then a stacked of Prototypical Representation Debiasing (PRD) modules are introduced to refine these prototypical representations iteratively obtain more precise support prototypical representation and query prototype for final mask prediction.
Rigorous experimentation and comprehensive analysis on three unique datasets indicate that our proposed method outstrips the current state-of-the-art techniques. This testifies to its formidable ability to generalize, along with its high reliability and sturdiness under diverse circumstances.

While the proposed method has shown promise, it does possess certain inherent limitations that warrant further investigation. First, the use of a self-supervised super-voxel technique for seen data annotation may inadvertently compromise segmentation accuracy to some degree, signifying a potential area for improvement through the incorporation of a more accurate automatic annotation method. Second, there's a need for a stronger pre-trained feature extractor to effectively boost its core generalization capability rather than overly focusing on feature alignment and refinement operations. Lastly, to construct a more discriminative prototype, it is highly recommended to extract and introduce supplementary information from the background region, thereby enriching the context and improving overall performance.

\ifCLASSOPTIONcaptionsoff
  \newpage
\fi
{
	\small
	 \bibliographystyle{IEEEtran}
\bibliography{reference}
}

\vspace{-3ex}
\begin{IEEEbiography}[{\includegraphics[width=1in,height=1.25in,clip,keepaspectratio]{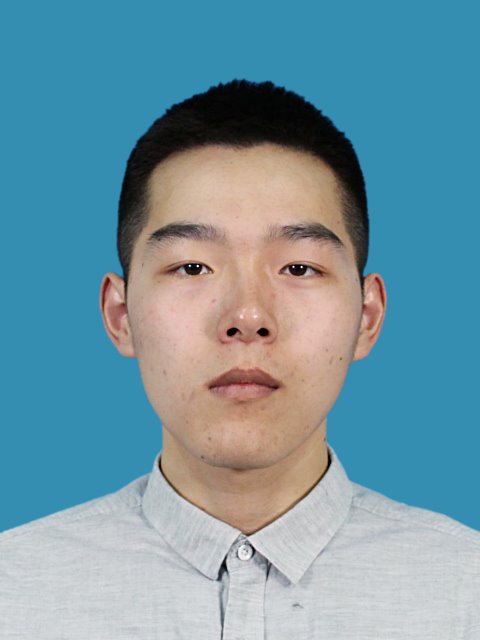}}]{Yazhou Zhu} received the B.Eng. degree from the Hua Loo-keng Honors College, Changzhou University, Changzhou, China, in 2018, and Master's degree in Software Engineering from Jiangnan University, Wuxi, China in 2021. Now he is currently working toward the Ph.D. degree in the School of Computer Science and Engineering, Nanjing University of Science and Technology, Nanjing, China. His current research interests include deep learning and medical image analysis.
\end{IEEEbiography}

\vspace{-3ex}
\begin{IEEEbiography}[{\includegraphics[width=1in,height=1.25in,clip,keepaspectratio]{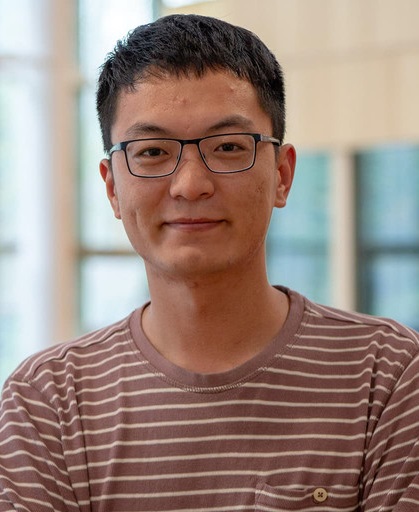}}]{Shidong Wang} is a Research Associate at NEOLab, School of Engineering, Newcastle University, UK. He received his PhD degree from the School of Computing Sciences, University of East Anglia (UEA), UK, in 2021. His research spans a breadth of domains including computer vision, deep learning, remote sensing, and environmental science, and he publishes in top-tier journals and conferences such as IEEE TPAMI, IJCV, ISPRS, TIP, AAAI and ACM MM. \end{IEEEbiography}

\vspace{-3ex}
\begin{IEEEbiography}[{\includegraphics[width=1in,height=1.25in,clip,keepaspectratio]{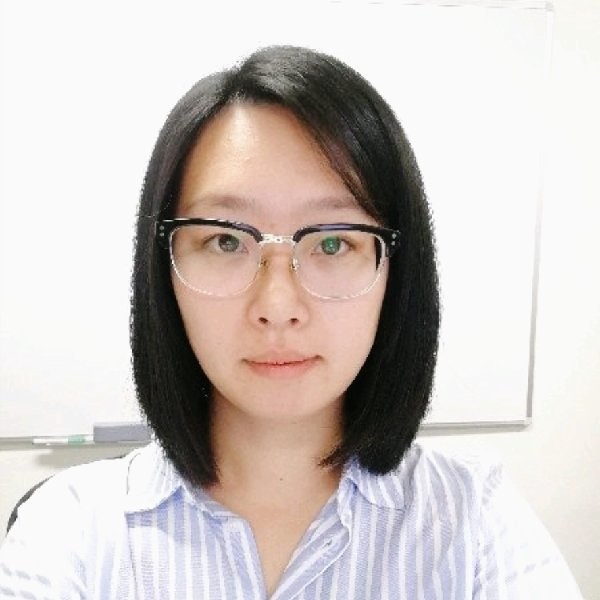}}]{Tong Xin} is a Lecturer in the School of Computing, Newcastle University. She obtained her PhD degree from the same department in 2020. Her interdisciplinary research focus on computer graphics, computer vision, medical image computing and data science. 
\end{IEEEbiography}

\vspace{-3ex}
\begin{IEEEbiography}[{\includegraphics[width=1in,height=1.25in,clip,keepaspectratio]{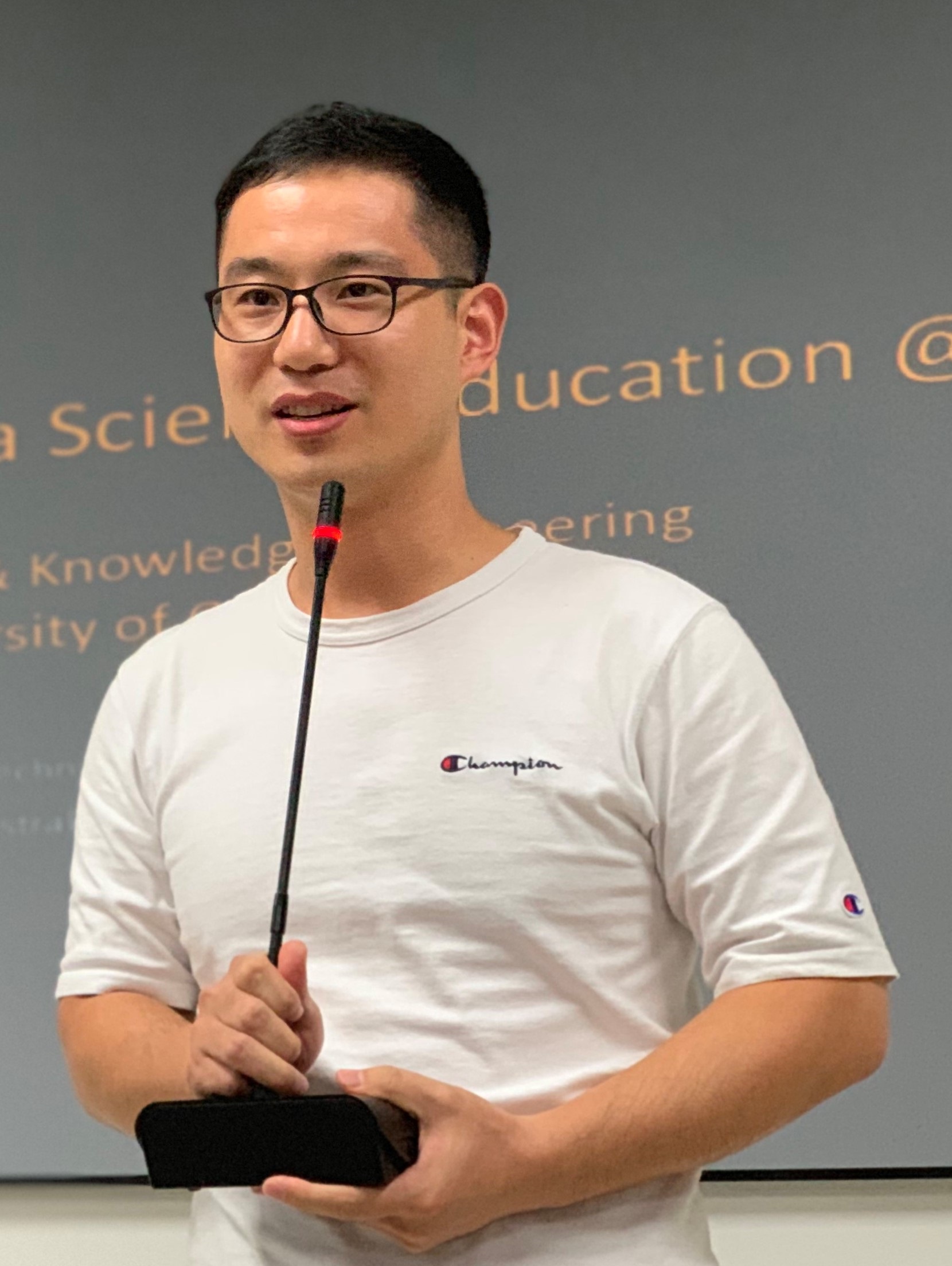}}]{Zheng Zhang} received his Ph.D. degree in Computer Applied Technology from the Harbin Institute of Technology. Since 2019, he has been with School of Computer Science \& Technology, Harbin Institute of Technology, Shenzhen, China, where he currently serves as the deputy director of the Shenzhen Key Laboratory of Visual Object Detection and Recognition, Shenzhen, China. His research interests mainly focus on multimedia content analysis and understanding, especially multimedia retrieval, multi-modal learning, and big data mining. He is currently at the Editorial Board of IEEE TAC, IEEE JBHI, and Information Fusion.
\end{IEEEbiography}

\vspace{-3ex}
\begin{IEEEbiography}[{\includegraphics[width=1in,height=1.25in,clip,keepaspectratio]{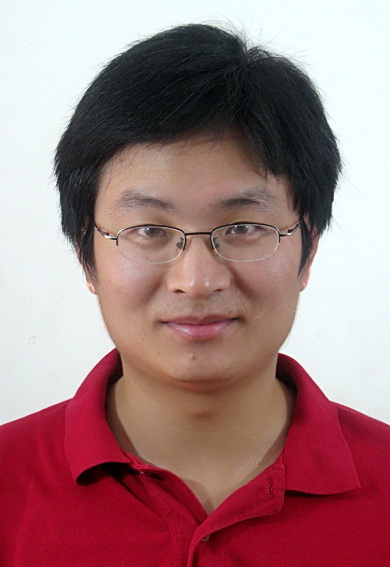}}]{Haofeng Zhang} is currently a Professor with the School of Computer Science and Engineering, Nanjing University of Science and Technology, China. He received the B.Eng. degree and the Ph.D. degree in 2003 and 2007 respectively from School of Computer Science and Technology, Nanjing University of Science and Technology, Nanjing, China. From Dec. 2016 to Dec. 2017, He was an academic visitor at University of East Anglia, Norwich, UK. His research interests include computer vision and mobile robot.
\end{IEEEbiography}

\vfill
% that's all folks
\end{document}